\documentclass{article}

\PassOptionsToPackage{numbers, compress}{natbib}
 \usepackage[preprint]{neurips_2026}

\usepackage{algorithm}  
\usepackage{algorithmic} 

\usepackage[utf8]{inputenc} 
\usepackage[T1]{fontenc}    
\usepackage{hyperref}       
\usepackage{url}            
\usepackage{booktabs}       
\usepackage{amsfonts}       
\usepackage{nicefrac}       
\usepackage{microtype}      
\usepackage{xcolor}         

\usepackage{microtype}
\usepackage{graphicx}
\usepackage{subcaption}
\usepackage{booktabs}

\usepackage{amsmath}
\usepackage{amssymb}
\usepackage{mathtools}
\usepackage{amsthm} 
\usepackage{enumitem}

\usepackage[capitalize,noabbrev]{cleveref} 
\def \w {\mathbf{w}}

\def \h {\mathbf{h}}  
\def \x {\mathbf{x}}

\def \th {\tilde{h}}  
\def \tH {\tilde{H}}  
\def \hh {\hat{h}}

\def \G {\mathcal{G}}  
\def \V {\mathcal{V}}

\def \E{\mathbb{E}}   
 
\def \N {\mathcal{N}}

\def \W {\mathbf{W}} 
\newcommand{\Mech}{\mathcal{M}}
\newcommand{\Xspace}{\mathcal{X}}
\newcommand{\Yspace}{\mathcal{Y}}
\newcommand{\eps}{\varepsilon} 

\newtheorem{theorem}{Theorem}[section]

\newtheorem{lemma}[theorem]{Lemma}

\newtheorem{definition}[theorem]{Definition}
\newtheorem{assumption}[theorem]{Assumption}
\newtheorem{remark}[theorem]{Remark} 

\title{Provably Communication-Efficient and Privacy-Preserving 
Federated Graph Neural Networks}

\author{\name Zhishuai Guo \email zguo@niu.edu\\
        \addr Northern Illinois University \\
        \name Wenhan Wu \email wwu25@charlotte.edu \\
        \addr University of North Carolina at Charlotte\\
        \name Chen Chen \email chen.chen@crcv.ucf.edu \\
        \addr University of Central Florida \\
        \name Lei Zhang \email zhanglei@niu.edu \\
        \addr Northern Illinois University\\
        \name Olivera Kotevska  \email kotevskao@ornl.gov\\
        \addr Oak Ridge National Laboratory \\
        \name Ravi K Madduri  \email madduri@anl.gov \\
        \addr Argonne National Laboratory}

\author{%
  Zhishuai Guo \\
  Department of Computer Science\\
  Northern Illinois University \\
  \texttt{zguo@niu.edu} \\
  \And   Wenhan Wu \\
  Department of Computer Science\\
  University of North Carolina at Charlotte \\
  \texttt{wwu25@charlotte.edu} \\
  \And  Chen Chen \\
  Department of Computer Science\\
  University of Central Florida \\
  \texttt{chen.chen@crcv.ucf.edu} \\
  \And  Lei Zhang \\
  Department of Computer Science\\
  Northern Illinois University \\
  \texttt{zhanglei@niu.edu} \\
  \And  Olivera Kotevska \\
  Computer Science and Mathematics Division\\
  Oak Ridge National Laboratory \\
  \texttt{kotevskao@ornl.gov} \\
  \And  Ravi K Madduri \\
  Data Science and Learning Division\\
  Argonne National Laboratory \\
  \texttt{madduri@anl.gov} \\
}

\begin{document}

\maketitle

\begin{abstract}

Graph neural networks (GNNs) achieve strong performance on relational data, but real-world graphs are often distributed across organizations that cannot share raw data due to privacy and policy constraints. Existing federated GNN methods either ignore cross-client links, leading to degraded accuracy, or require frequent embedding exchanges, incurring substantial communication and privacy costs.
We propose CE-FedGNN, a communication-efficient and privacy-preserving federated GNN framework for learning over such coupled graphs. Our approach avoids sharing raw data or per-round embeddings by infrequently exchanging aggregated node representations. To handle cross-client dependency and staleness, we introduce a moving-average estimator that continuously tracks node representations and enables their stable reuse across rounds.
To provide formal privacy guarantees for the released representations, we adopt the metric differential privacy (metric-DP) framework, which measures privacy with respect to distances in the learned embedding space rather than worst-case input perturbations. This yields meaningful guarantees at noise levels where standard differential privacy becomes overly conservative.
We establish convergence to a stationary point at a rate of $O(1/\sqrt{T})$ with $O(T^{3/4})$ communication complexity. In addition, we derive $(\eps,\delta)$-metric-DP guarantees via Rényi differential privacy composition under a public-cohort threat model. Experiments on synthetic interbank anti–money laundering benchmarks and citation networks demonstrate that CE-FedGNN achieves strong performance while significantly reducing communication and maintaining robustness under privacy-preserving noise.

\end{abstract}

\section{Introduction}
\begin{figure*}[htbp]
    \centering
    \begin{subfigure}{0.49\linewidth}
        \centering
        \includegraphics[width=\linewidth]{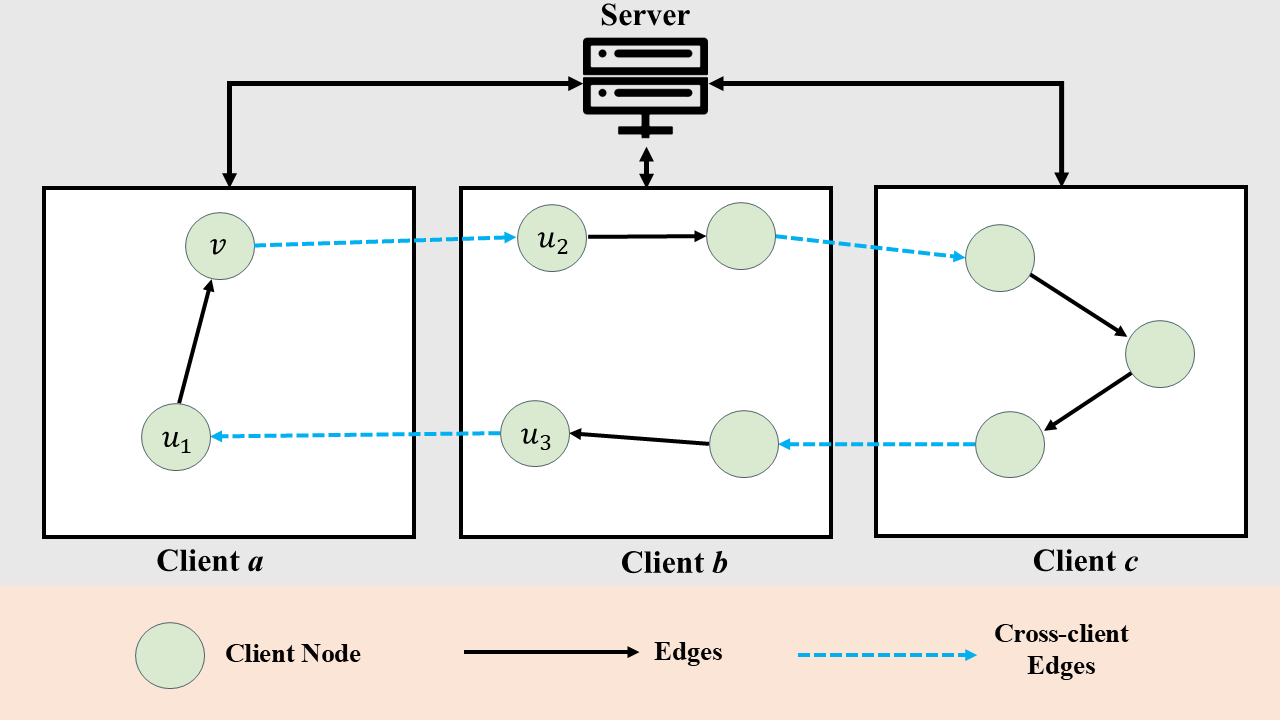}
        \caption[a]{Illustration of the challenge in FedGNN: a cycle that cannot be seen by either client alone.}
        \label{fig:introduction1}
    \end{subfigure}
    \hfill
    \begin{subfigure}{0.49\linewidth}
        \centering
        \includegraphics[width=\linewidth]{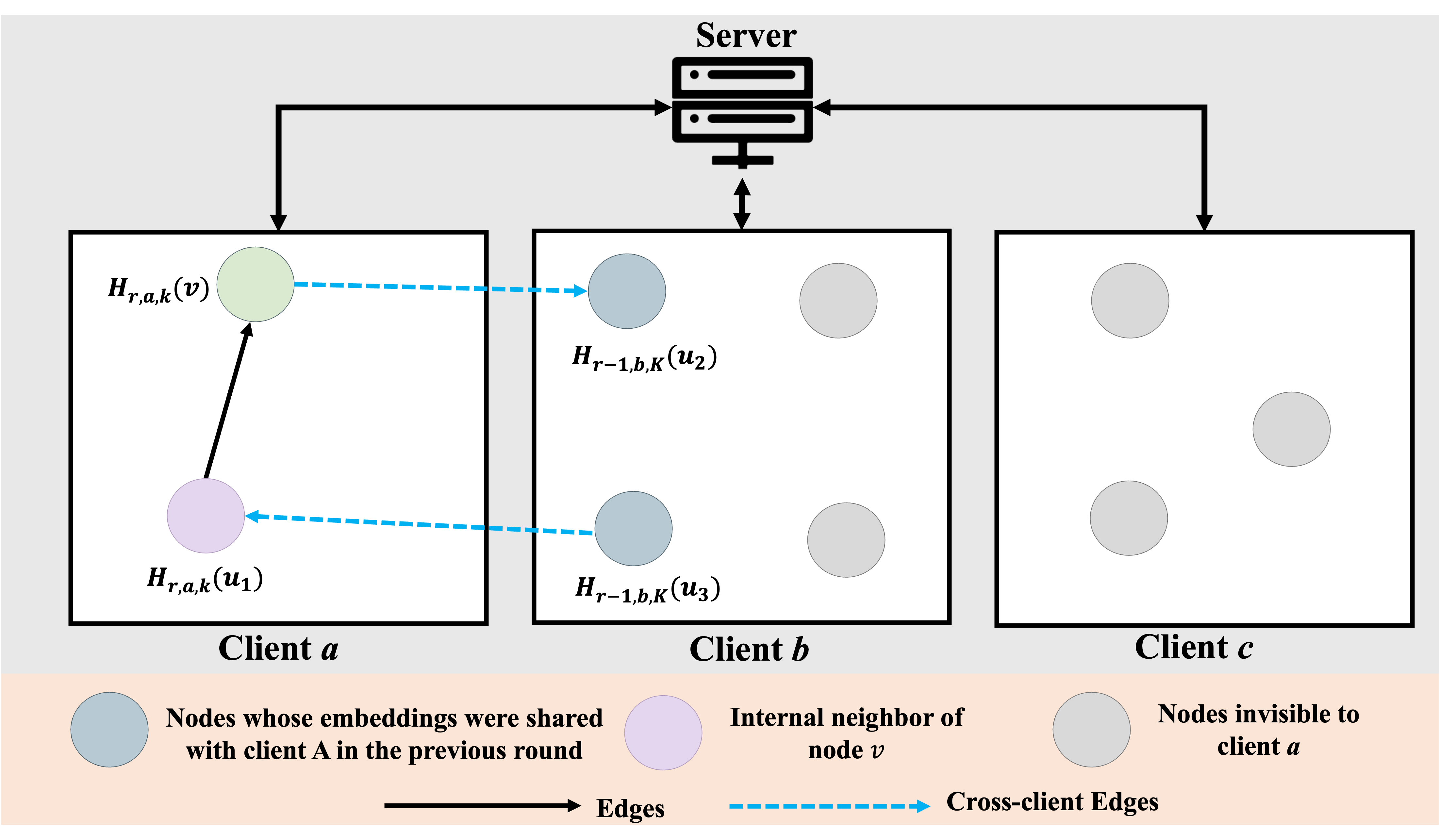}
        \caption{When the edge $(v,u_2)$ is sampled on client \textit{a}, it uses the recently shared embedding of node $u_2$ from \textit{b}.}
        \label{fig:introduction2}
    \end{subfigure}
    \caption{Illustration of the challenge in FedGNN and our algorithmic design.}
    \label{fig:introduction}
\end{figure*}

GNNs have emerged as a leading paradigm for learning from relational data in applications such as social network analysis, traffic forecasting, cybersecurity, and financial fraud detection \citep{shu2019beyond,derrow2021eta,lo2022graphsage,egressy2024provably}. By message passing over graph structures, GNNs capture relational patterns that are difficult for traditional models to represent. However, most existing GNN methods assume centralized access to a single, unified graph, an assumption often violated in practice. In many real-world settings, graph data are distributed across geographically and administratively separated locations within the same organization or across multiple organizations, making raw data aggregation impractical due to privacy regulations (e.g., GDPR, CCPA), security policies, and operational constraints. For example, in anti--money laundering detection, each financial institution observes only a local subgraph of the global transaction network, which is insufficient to detect global patterns (Figure~\ref{fig:introduction}(a)).

FL provides a natural framework for collaborative model training without centralizing data. However, standard FL methods such as FedAvg~\citep{konevcny2016federated,mcmahan2017communication} are designed for settings with independent local datasets, where the global objective decomposes as an average of client-specific losses. Graph-structured data fundamentally violates this assumption: edges may connect nodes owned by different clients. As a result, neighborhood aggregation, which is the core GNN operation, depends on information that is not locally available. Ignoring cross-client edges yields incomplete and biased representations, while naively exchanging node embeddings at every iteration incurs prohibitive communication overhead. These difficulties are further exacerbated by the multi-layer compositional structure of GNNs.

Existing federated GNN methods face a fundamental trade-off between modeling accuracy and communication efficiency: they either ignore cross-client edges and lose global structure, exchange embeddings at every iteration with prohibitive overhead, or share embeddings infrequently in ways that fail to adapt to representation drift~\citep{he2021fedgraphnn,hu2022fedgcn,wu2022federated,yao2023fedgcn,qiuswift,zhang2021subgraph}. None of these approaches simultaneously achieves accurate cross-client modeling, low communication cost, and scalability.

Privacy considerations compound the problem. Although FL avoids raw data sharing, exchanged embeddings can still leak sensitive information~\citep{duddu2020quantifying,li2020adversarial,zhang2022model,zhang2024survey}, motivating the addition of calibrated noise. 
While recent work has provided differential privacy (DP) guarantees for centralized GNN training~\citep{daigavane2021node,sajadmanesh2023gap}, these methods protect the released model and do not address per-node embedding releases exchanged during federated training. For embedding release, the worst-case $L_2$ sensitivity is the embedding-ball diameter, rendering standard DP vacuous at practical noise levels.


To address these challenges jointly, we propose a communication-efficient and privacy-preserving federated GNN framework. Our method explicitly models cross-client graph coupling while infrequently exchanging aggregated node embeddings, and it adopts the metric-DP framework~\citep{chatzikokolakis2013broadening} to characterize privacy by $L_2$ distance on the trained embedding space rather than worst-case input adjacency. This yields meaningful guarantees at moderate noise levels, where standard DP would give vacuous bounds. We provide formal analyses of both convergence and privacy.

\textbf{Our contributions are summarized as follows.}
\begin{itemize}[leftmargin=*]
\item \textbf{Decomposition framework for federated GNNs.}
We propose a tailored decomposition framework that maintains moving-average estimators of node embeddings to reduce variance and bias in mini-batch training. Crucially, this mechanism is applied only to nodes---not edges---which is important in practice since the number of edges can vastly exceed the number of nodes. When a neighbor belongs to another client, we reuse its most recently shared moving-average embedding (Figure~\ref{fig:introduction}(b)), and explicitly account for the resulting latency error in our analysis. This design restricts cross-client interactions to one-hop neighbors and eliminates multi-hop cross-client sampling. With $T$ iterations, our method achieves an $O(1/\sqrt{T})$ convergence rate to a stationary point while requiring only $O(T^{3/4})$ communication rounds.

\item \textbf{Metric-DP guarantee for federated GNN embedding release.}
We adopt the metric-DP framework to formally protect released node embeddings, calibrating Gaussian noise to $L_2$ distance on the trained embedding space. We provide $(\eps,\delta)$-metric-DP composition guarantees over communication rounds via Rényi DP, accounting for the public-cohort threat model in federated GNN protocols where subsampling amplification does not apply. To our knowledge, this is the first application of metric-DP to federated GNN training. We also analyze the impact of injected noise on convergence dynamics, showing graceful degradation in optimization quality.

\item \textbf{Empirical evaluation.}
We conduct extensive experiments on synthetic anti--money laundering benchmarks and real-world citation networks. CE-FedGNN consistently outperforms federated GNN baselines while using substantially fewer communication rounds, and retains strong utility under noise levels that translate to meaningful metric-DP guarantees.
\end{itemize}

\vspace{-0.15in}
\section{Related Work}
\textbf{Graph Neural Networks.}
Many real-world problems involve entities connected by complex relational structures that cannot be naturally represented in Euclidean space. GNNs provide a principled framework for learning from such data by propagating and aggregating information over graph topology, achieving strong empirical performance across domains including social networks~\citep{badrinath2025omnisage,hou2025optimize}, transportation systems~\citep{derrow2021eta,zheng2023spatio}, physical simulation~\citep{pmlr-v119-sanchez-gonzalez20a,jain2025latticegraphnet}, biological and molecular modeling~\citep{bongini2021molecular}, combinatorial optimization~\citep{dudzik2022graph,bevilacqua2023neural}, and financial fraud detection~\citep{egressy2024provably,lin2024fraudgt}. A broader overview of recent advances in GNN architectures is provided in Appendix~\ref{appx:GNN}.

\textbf{Communication-Efficient Federated Learning.}
Communication efficiency is a central challenge in FL. A large body of work studies this problem under the assumption that the global objective decomposes as an average of independent client losses, leading to communication-efficient algorithms based on periodic averaging, variance reduction, or local updates~\citep{konevcny2016federated,mcmahan2017communication,stich2018local,yu2019linear,yu2019parallel,yang2013trading,karimireddy2020scaffold,kairouz2021advances,khaled2020tighter,woodworth2020minibatch,woodworth2020local,haddadpour2019local,deng2021local,deng2020distributionally,liu2020decentralized,sharma2022federated,li2022communication,huang2022federated,pmlr-v162-tarzanagh22a}.
These methods, however, are not directly applicable to settings with \emph{coupled objectives}, where client data are interdependent. Several works address specific coupling structures: \citet{yuan2021federated,guo2020communication} study federated AUC maximization via minimax reformulations, while \citet{gao2022convergence} analyze compositional objectives without cross-client data dependencies. \citet{guo2023fedxl} propose a communication-efficient framework for general pairwise objectives via active--passive gradient decomposition, but the approach does not extend to federated GNNs with multi-layer compositional dependencies between node representations and model parameters. Overall, existing approaches either rely on problem-specific reformulations or heuristic designs and do not provide convergence guarantees for coupled graph-structured objectives~\citep{https://doi.org/10.48550/arxiv.2207.09158,DBLP:journals/corr/abs-2010-08982,wu2022federated,li2022fedgrec}.

\textbf{Federated Graph Neural Networks.}
Federated GNNs introduce additional challenges due to edges that span multiple clients. Existing approaches either ignore such edges to simplify training~\citep{du2022malicious,peng2022domain,scardapane2020distributed,he2021fedgraphnn,baek2023personalized}, or explicitly model them by exchanging node embeddings at every iteration~\citep{wu2022federated,wu2021fedgnn}. 
Several intermediate methods attempt to reduce communication: \citet{yao2023fedgcn} share embeddings only at initialization, \citet{qiuswift} exploit the global graph infrequently, and \citet{zhang2021subgraph} generate missing nodes locally. However, these approaches often fail to adapt to representation drift, obscure global structural patterns, or rely on complex procedures that limit scalability. 
\citet{aliakbari2025subgraph} share global connectivity via spectral methods, but incur expensive matrix decompositions and assume static graphs, limiting applicability to dynamic settings.

\textbf{Privacy in FL and GNNs.}
Although FL avoids direct sharing of raw data, privacy risks arise through the exchange of model parameters, gradients, or intermediate representations~\citep{zhu2019deep,duddu2020quantifying}.
A common mitigation is to inject randomized perturbations under the differential privacy (DP) framework~\citep{dwork2014algorithmic,zhu2019deep, abadi2016deep, mcmahan2018learning, truex2020ldp, wei2020federated}.
While DP has been extensively studied in standard FL settings, most federated GNN methods do not explicitly account for privacy risks arising from the exchange of intermediate representations.
Although aggregated embeddings are generally considered less sensitive than raw features~\citep{yao2023fedgcn}, intermediate GNN embeddings can still leak information~\citep{duddu2020quantifying, li2020adversarial, zhang2022model, zhang2024survey}.

Applying standard DP directly to per-node embedding release is impractical: the worst-case $L_2$ sensitivity over neighbors equals the diameter of the embedding ball, forcing noise scales that destroy utility. We instead adopt the metric differential privacy (metric-DP) framework of \citet{chatzikokolakis2013broadening}, which parameterizes privacy by a distance on the input space, capturing indistinguishability within a semantic neighborhood rather than against worst-case alternatives. Metric-DP has been applied to single-release embedding privacy in NLP~\citep{feyisetan2020privacy,xu2020differentially,bollegala2025metric}. To our knowledge, metric-DP has not previously been applied to federated GNN training, where multi-round composition and per-node release introduce considerations beyond the single-release case. We extend the framework via Rényi DP composition~\citep{mironov2017renyi, mironov2019r}, with the formal mechanism deferred to Section \ref{sec:dp:noise} and Appendix~\ref{app:privacy}.

\vspace{-0.1in}   
\section{Method}
\label{sec:method}
We introduce a communication-efficient federated GNN algorithm and extend it with a metric-DP mechanism that protects released node embeddings during cross-client communication. 

\subsection{Problem Statement}
\label{sec:statement}

Many modern GNNs follow the message-passing framework~\citep{wu2020comprehensive}, in which node representations are updated by aggregating information from their neighbors. For a graph $\G=(\V,\mathcal{E})$, layer $l$ updates each node $v\in\V$ as
\begin{equation}
\h^{(l)}(v) = \text{UPDATE}^{(l)}\!\Big(\h^{(l-1)}(v),\, \text{AGGREGATE}^{(l)}\big(\{\h^{(l-1)}(u) : u \in \mathcal{N}(v)\}\big)\Big),
\end{equation}
where $\mathcal{N}(v)$ is the neighbor set of $v$ and $\text{AGGREGATE}^{(l)}$ is permutation-invariant. We focus on a common class of message-passing GNNs whose aggregation follows GCN~\citep{kipf2017semi}, GraphSAGE-mean~\citep{hamilton2017inductive}, or GIN~\citep{xu2018powerful}. For clarity, we use the GraphSAGE-mean form,
\begin{equation}
\small
\h^{(l)}(v) = \phi\!\bigg(\mathbf{W}^{(l)} \cdot \frac{1}{|\mathcal{N}(v)|}\sum_{u \in \mathcal{N}(v)} \h^{(l-1)}(u)\bigg),
\end{equation}
where $\phi(\cdot)$ is a nonlinear activation. We denote the aggregated message by $\hh^{(l)}(v) := \frac{1}{|\mathcal{N}(v)|} \sum_{u \in \mathcal{N}(v)} \h^{(l-1)}(u)$ and the pre-activation by $\th^{(l)}(v) := \mathbf{W}^{(l)} \hh^{(l)}(v)$, so $\h^{(l)}(v) = \phi(\th^{(l)}(v))$. Corresponding GCN and GIN formulations are in Appendix~\ref{appx:GNN}.

For node- or edge-level prediction, a task-specific head is applied to the final-layer embeddings: $\hat{y}_x = f(\mathbf{W}^{L+1}; \h(x))$, $\mathcal{L}(x) = \ell(\hat{y}_x, y_x)$, $x \in \mathcal{X}$, where $\mathcal{X}=\V$ for node tasks and $\mathcal{X}=\mathcal{E}$ for edge tasks. For edge prediction on $e=(u,v)$, an edge representation is computed by aggregating endpoint embeddings:
\begin{equation}
\small
\h(e) = \phi\!\left(\mathbf{W}^e \cdot \frac{\h^{(L)}(u) + \h^{(L)}(v)}{2}\right).
\end{equation} 

Let $\mathbf{W}$ denote all model parameters. The global objective is
$F(\mathbf{W}) = \frac{1}{N} \sum_{i=1}^N F_i(\mathbf{W})$ with $F_i(\mathbf{W}) = \frac{1}{|\mathcal{X}_i|} \sum_{x \in \mathcal{X}_i} \mathcal{L}(x)$,
where $\mathcal{X}_i$ is the training set on client $i$. 

We consider a realistic deployment in which an edge connecting two clients is observable to both parties (i.e., edge attributes are shared), while node attributes remain private. For example, when a transaction occurs between two banks, both institutions observe the transaction attributes (time, amount), but each retains only its own node features.

\vspace{-0.1in}
\subsection{A Communication-Efficient Algorithm}
\label{sec:algorithm}
To illustrate the challenge posed by cross-client edges, consider an edge $e=(u,v)$ on client $i$ where $v$ is hosted by another client. The gradient of the local objective with respect to the edge-head parameters $\mathbf{W}^e$ is
\begin{equation}
\small
\frac{\partial F_i(\mathbf{W};e)}{\partial \mathbf{W}^e}
=
\frac{\partial \mathcal{L}(e)}{\partial h(e)}\,
\phi'\!\left(\mathbf{W}^e \cdot \frac{\h^{(L)}(u) + \colorbox{blue!30}{$\h^{(L)}(v)$}}{2}\right)
\cdot \frac{\h^{(L)}(u) + \colorbox{blue!30}{$\h^{(L)}(v)$}}{2},
\end{equation}
where the highlighted term is the embedding of a cross-client neighbor. This dependence creates four challenges:
(i) \emph{cross-client dependency} ($\h^{(L)}(v)$ is unavailable on client $i$);
(ii) \emph{communication cost} (naive per-iteration exchange is prohibitive);
(iii) \emph{gradient backpropagation} (chain-rule terms involving cross-client embeddings cannot be differentiated locally); and
(iv) \emph{stochastic gradient bias} (neighborhood subsampling and partial cross-client information compound bias under nonlinear message passing). 

\textbf{Moving-average estimators.}
We address the above challenges by maintaining moving-average estimators of (i) intermediate node embeddings and (ii) stochastic gradients, which together provide low-variance approximations under minibatch training. Let the subscript $(r,k,i)$ denote local iteration $k$ of round $r$ on client $i$. The moving-average forward update at layer $l$ is
\begin{equation}
\small
\begin{split}
\tH^{(l)}_{r,k,i}(v)
&= (1-\gamma)\,\tH^{(l)}_{r,k-1,i}(v) + \gamma\,
\mathbf{W}^{(l)}_{r,k-1,i} \cdot
\frac{1}{|\mathcal{N}_{r,k,i}(v)|}
\sum_{u \in \mathcal{N}_{r,k,i}(v)} H^{(l-1)}_{r,k,i}(u), \\
H^{(l)}_{r,k,i}(v) &= \phi\!\left(\tH^{(l)}_{r,k,i}(v)\right),
\end{split}
\label{eq:forward_node}
\end{equation}
where $\mathcal{N}_{r,k,i}(v)$ is the sampled neighbor set of $v$ on client $i$ at iteration $(r,k)$, and $H^{(0)}(\cdot)=\h^{(0)}(\cdot)$. We also define the virtual global average $\mathbf{W}_{r,k} := \frac{1}{N}\sum_{i=1}^N \mathbf{W}_{r,k,i}$, which is not explicitly computed during training.

We make the following assumption for the analysis. 

\begin{assumption}
\label{ass:1}
~
\begin{itemize} 
    \item[(i)] $f$, and $h$ are $C_0$-Lipschitz and $C_1$-smooth in $\W$, and $\phi(\cdot)$ is $C_0$-Lipschitz and $C_1$-smooth.
    \item[(ii)] The stochastic gradient estimator satisfies $\E\|\hat{\nabla} F_i^h(\w;B) - \nabla F_i(\w)\|^2 \leq \sigma^2$, and the gradient estimator based on embedding estimators is bounded as $\|\hat{\nabla} F_i^H(\w;B)\|^2 \leq D^2$, where the superscripts $h$ and $H$ denote computation using true embeddings and moving-average estimators, respectively.
    \item[(iii)] $F$ is bounded below: there exists $\W_*$ with $F(\W) \geq F(\W_*) > -\infty$.
    \item[(iv)] Model parameters and embeddings are bounded: $\|\W\|^2 \leq C_W^2$ and $\|H(\cdot)\|^2 \leq C_H^2$.
\end{itemize}
\end{assumption}
The assumptions are standard in the optimization literature, with the smoothness assumption of (graph) neural networks justified in \citep{awasthi2021convergence,allen2019convergence}.
The behavior of the embedding estimator is characterized by the following result.
\begin{lemma}
\label{lem:H_estimate}
Under Assumption~\ref{ass:1}, let $p$ denote the minimum probability that a node is sampled in a local minibatch. Then Algorithm~\ref{alg:ce_fedgnn} ensures
\begin{equation}
\small
\frac{1}{R K} \sum_{r=1}^{R} \sum_{k=1}^{K}
\mathbb{E}\!\left[\| H^{(l)}_{r,k,i}(u) - h^{(l)}(u) \|^2\right]
\le O\!\left(
\frac{1}{\gamma p R K} + \frac{\gamma}{p} + \gamma^2 K^2
+ \frac{\|\mathbf{W}_{r,k} - \mathbf{W}_{r,k-1}\|^2}{\gamma^2}
\right).
\end{equation}
\end{lemma}

\noindent\textbf{Remark.}
With appropriate choices of $\gamma$ and $K$, the first three terms decrease as $R$ grows. The final term reflects model drift across local updates and is controlled by ensuring that updates use low-variance gradient estimates (handled next). As a result, only one-hop node embeddings need to be exchanged across clients: each node maintains a moving-average estimate of its representation, removing the need for multi-hop cross-client sampling.

\textbf{Cross-client gradient handling.}
The stochastic gradient with respect to $\mathbf{W}^e$ on client $i$ is
\begin{equation}
\small
\hat{\nabla} F_i(\mathbf{W}^{e}_{r,k,i}; B_{r,k,i})
= \frac{\partial \mathcal{L}(e)}{\partial h(e)}\,
\phi'\!\left(\mathbf{W}^e_{r,k,i} \cdot \hat{h}_{r,k,i}(e)\right)\,\hat{h}_{r,k,i}(e),
\end{equation}
with
\begin{equation}
\small
\hat{h}_{r,k,i}(e) = \frac{1}{2}\left(\h^{(L)}_{r,k,i}(u) + \colorbox{blue!30}{$\h^{(L)}_{r-1,K,c(v)}(v)$}\right),
\end{equation}
where $c(v)$ denotes the host client of $v$. The estimator $\hat{h}_{r,k,i}(e)$ reuses the most recently shared embedding of cross-client node $v$, which is generally stale. By the chain rule, computing the full gradient would require differentiating through $\h^{(L)}(v)$, which is not locally available.

We exploit the fact that the edge $e=(u,v)$ is observed by both incident clients $c(u)$ and $c(v)$. When the edge is sampled on client $c(u)$, that client computes the gradient component involving the embedding of its locally owned node $u$, treating the cross-client embedding $\h^{(L)}(v)$ as a fixed input via its most recently shared moving-average value. Symmetrically, when the edge is sampled on client $c(v)$, that client computes the component involving $v$'s embedding. Since each client backpropagates through only its own endpoint, neither computes the full gradient locally. However, when edge sampling is properly weighted across the two incident clients (e.g., equal sampling probabilities or
local-gradient reweighting), the aggregated update
$\frac{1}{N}\sum_i \hat{\nabla} F_i(\mathbf{W}^{e}_{r,k,i}; B_{r,k,i})$
recovers an unbiased estimate of the full gradient with respect to
$\mathbf{W}^e$.

We further stabilize optimization with a moving-average gradient estimator:
\begin{equation}
\small
G^{(l)}_{r,k,i} = (1-\beta)\,G^{(l)}_{r,k-1,i} + \beta\,\hat{\nabla} F_i(\mathbf{W}_{r,k-1,i}; B_{r,k,i}), \beta \in (0,1]. 
\label{eq:G_ma}
\end{equation}

The behavior of the gradient estimator is characterized by:
\begin{lemma}
\label{lem:G_estimate}
Under Assumption~\ref{ass:1}, let $\bar{G}_{r,k} := \frac{1}{N} \sum_{i=1}^N G_{r,k,i}$. Then Algorithm~\ref{alg:ce_fedgnn} ensures
\begin{equation*}
\small
\E\!\left[\|\bar{G}_{r,k} - \nabla F(\mathbf{W}_{r,k})\|^2\right]
\le O\!\bigg(\frac{1}{\beta R K} + \frac{1}{\gamma p R K} + \frac{\beta}{N} + \frac{\gamma}{p}
+ \beta^2 K^2 + \frac{\eta^2}{\beta^2}\|\nabla F(\mathbf{W}_{r,k-1})\|^2\bigg).
\end{equation*}
\end{lemma}

\begin{algorithm}[t] 
\caption{CE-FedGNN}
\begin{algorithmic}[1] 
\STATE \textbf{On Server:}
\STATE Initialize global model $\{\mathbf{W}^{(l)}\}_{l=1}^{L+1}$ and global embedding buffer $\mathcal{H}$.
\FOR{$r = 1, 2, \ldots, R$}
    \STATE \textbf{Clients execute in parallel:}
    \STATE \quad $(\mathbf{W}_{r,K,i}, G_{r,K,i}, \mathcal{H}_{r,i}) \leftarrow \textsc{LocalUpdate}(\mathbf{W}_{r,0,i}, \mathcal{H})$
    \STATE $\mathbf{W}_{r} = \frac{1}{N}\sum_{i \in [N]} \mathbf{W}_{r,K,i}$,\quad $G_{r} = \frac{1}{N}\sum_{i \in [N]} G_{r,K,i}$
    \STATE $\mathbf{W}_{r+1,0, i} = \mathbf{W}_{r}$,\quad $G_{r+1,0,i} = G_{r}$
    \STATE Update $\mathcal{H}$ using $\{\mathcal{H}_{r,i}\}_{i=1}^N$.
\ENDFOR
\STATE \textbf{Output:} Final global model $\mathbf{W}_R$.
\STATE \rule{\linewidth}{0.4pt}
\STATE \textbf{LocalUpdate}$(\mathbf{W}_{r,0,i}, \mathcal{H})$ \textbf{on client $i$:}
\STATE Receive current model $\mathbf{W}_{r,0,i}$ and required cross-client embeddings from $\mathcal{H}$.
\FOR{$k = 1, 2, \ldots, K$}
    \STATE Sample minibatch $\mathcal{B}_{r,k,i}$ of seed nodes/edges, then sample up to $L$-hop neighborhoods.
    \STATE Update node embeddings via the moving-average forward rule~\eqref{eq:forward_node}.
    \STATE Update gradient estimator via~\eqref{eq:G_ma}.
    \STATE Update model: $\mathbf{W}_{r,k,i} = \mathbf{W}_{r,k-1,i} - \eta G_{r,k,i}$.
\ENDFOR
\STATE Return $(\mathbf{W}_{r,K,i}, G_{r,K,i}, \mathcal{H}_{r,i})$ to the server.
\end{algorithmic}
\label{alg:ce_fedgnn}
\end{algorithm}
      
\vspace{-0.1in}
\noindent\textbf{Remark.}
With appropriate choices of $\beta$, $\gamma$, and $K$, the variance of $G$ decreases over rounds, and the estimation error diminishes near a stationary point. 
Combining Lemmas~\ref{lem:H_estimate} and~\ref{lem:G_estimate} yields the following convergence guarantee.
\begin{theorem}
\label{thm}
Under Assumption~\ref{ass:1}, Algorithm~\ref{alg:ce_fedgnn} satisfies
\begin{equation*}
\small
\frac{1}{R} \sum_{r=1}^{R}\E\!\left[\|\nabla F(\mathbf{W}_{r-1})\|^2\right]
\le O\!\bigg(\frac{1}{\eta R K} + \frac{1}{\beta R K} + \frac{1}{\gamma p R K}
+ \frac{\beta}{N} + \frac{\gamma}{p} + \beta^2 K^2\bigg).
\end{equation*}
\end{theorem} 

\noindent\textbf{Remark.}
With $\gamma = \Theta(R^{-2/3})$, $\beta = \Theta(R^{-2/3})$, $\eta = \Theta(R^{-2/3})$, and $K = \Theta(R^{1/3})$, Theorem~\ref{thm} yields $\frac{1}{R}\sum_r \E\|\nabla F(\mathbf{W}_{r-1})\|^2 = O(R^{-2/3})$. Equivalently, with $T = RK$ total local iterations and $R = \Theta(T^{3/4})$, we obtain an $O(T^{-1/2})$ convergence rate. Reaching $\frac{1}{R}\sum_r \E\|\nabla F(\mathbf{W}_{r-1})\|^2 \le \epsilon^2$ requires $R = O(\epsilon^{-3})$, $K = O(\epsilon^{-1})$, and $\eta = \Theta(\beta) = \Theta(\gamma) = O(\epsilon^2)$, giving iteration complexity $O(\epsilon^{-4})$ and communication complexity $O(\epsilon^{-3})$.

The dependence on the number of clients $N$ appears via $\beta/N$, indicating that more participating clients reduce gradient estimation error. When $p$ scales with $N$ (e.g., even data partitioning across more clients), this can yield linear speedup in $N$: with $R = O(\epsilon^{-3})$, $K = O((N\epsilon)^{-1})$, and $\eta = \Theta(\beta) = \Theta(\gamma) = O(N\epsilon^2)$, the total iteration complexity is $T = RK = O(N^{-1}\epsilon^{-4})$.

At the end of each round, clients send updated model parameters, gradient estimators, and boundary-node embeddings to the server, which routes embeddings to neighboring clients. Each client communicates only updated boundary-node embeddings. When an embedding is not updated, the most recently shared value is reused, with the moving-average mechanism ensuring accuracy in expectation.

\vspace{-0.1in}
\subsection{Privacy-Preserving Extension via Metric-DP}
\label{sec:privacy}
\label{sec:dp:noise}

To formally protect released node embeddings, we extend the framework with calibrated Gaussian noise at communication time and analyze the resulting privacy and convergence guarantees.

\textbf{Threat model.}
We consider an honest-but-curious adversary that observes all messages exchanged between clients during federated training, including the identity of each round's released cohort. Crucially, in our protocol, the per-round update set is publicly observable, since clients explicitly 
broadcast updated boundary-node embeddings. We do not invoke 
privacy amplification by Poisson subsampling~\citep{mironov2019r}: 
the adversary's uncertainty about which records influenced a given 
release is zero.

\textbf{Mechanism.}
During local updates, the embedding estimator $H$ and gradient estimator $G$ remain unchanged. Gaussian noise is injected only when these quantities and model parameters are communicated across clients. For cross-client edges, the shared embedding is perturbed and the forward computation becomes
\begin{equation}
\small
\hat{h}_{r,k,i}(e) = \frac{1}{2}\bigg(\h^{(L)}_{r,k,i}(u) + \colorbox{blue!30}{$\h^{(L)}_{r-1,K,c(v)}(v) + \mathcal{N}(0,\sigma_0^2 I)$}\bigg).
\end{equation}
Local updates of $\mathbf{W}$ and $G$ are unchanged, but their aggregated forms at communication rounds are perturbed:
\begin{equation}
\small
\mathbf{W}_{r+1,0,i} = \frac{1}{N}\sum_{j=1}^N \mathbf{W}_{r,K,j} + \mathcal{N}(0,\sigma_1^2 I),
\quad
G_{r+1,0,i} = \frac{1}{N}\sum_{j=1}^N G_{r,K,j} + \mathcal{N}(0,\sigma_2^2 I).
\end{equation}

\textbf{Metric-DP guarantee.}
We adopt metric differential privacy~\citep{chatzikokolakis2013broadening}, which generalizes standard DP by parameterizing the privacy guarantee by a distance metric on the input space.

\begin{definition}[Metric Differential Privacy]
\label{def:metricdp}
A mechanism $\Mech : \Xspace \to \Yspace$ satisfies $(\eps, \delta)$-metric-DP under metric $d$ if for all $x, x' \in \Xspace$ and all measurable $S$:
\[
\Pr[\Mech(x) \in S] \leq e^{\eps \cdot d(x, x')} \cdot \Pr[\Mech(x') \in S] + \delta.
\]
Standard $(\eps,\delta)$-DP corresponds to $d$ being the discrete metric on neighboring datasets.
\end{definition}

\begin{theorem}[Composed Gaussian mechanism for metric-DP]
\label{thm:mdp_composition}
For an $R'$-fold composition of the Gaussian mechanism, where $R'$ denotes the number of rounds a node's embeddings are shared, applying Gaussian noise with standard deviation $\sigma_0$ to $L_2$-normalized embeddings under public-cohort release satisfies $(\eps(\rho), \delta)$-metric-DP under $L_2$ distance at any $\rho > 0$, with
\begin{equation}
\small
\eps(\rho) = \min_{\alpha > 1}\left[\,\frac{R' \alpha \rho^2}{2 \sigma_0^2} + \log\!\frac{\alpha-1}{\alpha} - \frac{\log(\delta\alpha)}{\alpha-1}\right].
\end{equation}
\end{theorem}

\textbf{Remark.} Setting $\rho$ to be the diameter of the unit $L_2$ sphere of the embedding in Theorem~\ref{thm:mdp_composition} recovers the standard $(\eps,\delta)$-DP guarantee for the composed Gaussian mechanism without subsampling, as in Abadi et al.~\citep{abadi2016deep} for the no-subsampling regime. 
It is noticeable that $R'\ll R \ll T$ since each round only uses a subset of all nodes and our infrequent communication approach reduces $R$.  

The proof of Theorem~\ref{thm:mdp_composition} and a discussion of the public-cohort threat model are deferred to Appendix~\ref{app:privacy}. Implementation uses the Rényi DP accountant with noise multiplier $z=\sigma_0/\rho$ and sample rate $1$ (no subsampling amplification).

\textbf{Choice of $\rho$.}
The metric-DP framework parameterizes privacy by a distance $\rho$, which we instantiate via a $k$-anonymity-style construction: each account is required to be indistinguishable from its $k$ nearest neighbors in the trained embedding space. Let $d_k(v)$ denote the $L_2$ distance from $v$ to its $k$-th nearest neighbor among $L_2$-normalized released embeddings. We set $\rho_{\max} = Q_q(\{d_k(v) : v \in V\})$, the $q$-th percentile of the $k$-th nearest neighbor distance distribution. The resulting privacy claim states that, for at least $q\%$ of accounts, the released embedding is $(\eps,\delta)$-metric-DP indistinguishable from each of its $k$ nearest neighbors.

\textbf{Convergence under noise.}
The effect of noise on the embedding and gradient estimators is bounded by Lemmas~\ref{lem:dp_H_estimate} and~\ref{lem:dp_G_estimate} in Appendix~\ref{app:privacy}. The overall convergence is Theorem \ref{thm:dp} in Appendix \ref{app:privacy}.
\begin{theorem}
\label{thm:dp}
Under Assumption~\ref{ass:1}, with $\eta=\Theta(\beta)=\Theta(\gamma)$, the noise-perturbed version of Algorithm~\ref{alg:ce_fedgnn} satisfies
\begin{equation*}
\small
\frac{1}{R}\sum_{r=1}^R \E\!\left[\|\nabla F(\mathbf{W}_{r-1})\|^2\right]
\le O\!\Big(\frac{1}{\eta R K} + \beta + \beta^2 K^2 + \sigma_0^2 + \frac{\sigma_1+\sigma_2}{\beta}\Big).
\end{equation*}
\end{theorem}

\noindent\textbf{Remark.}
The convergence bound depends on the noise levels of all communicated quantities. In our setting, model parameters $\mathbf{W}$ and gradient estimators $G$ aggregate information from many nodes and edges within a batch, so relatively small noise standard deviation $\sigma_1, \sigma_2$ suffices for typical privacy goals on these quantities. Released embeddings  are the primary target of our metric-DP analysis: the bound exhibits a milder dependence on $\sigma_0$ (entering quadratically) than on $\sigma_1$ and $\sigma_2$ (entering as $1/\beta$), suggesting that stronger embedding perturbation can be tolerated. This favorable dependence aligns with the design of metric-DP, which yields meaningful privacy guarantees at moderate $\sigma_0$ values where standard DP is vacuous. 

\section{Experiments}
\label{sec:experiments}
We evaluate CE-FedGNN on three axes: predictive performance against federated GNN baselines, communication efficiency under varying communication intervals, and privacy--utility tradeoff under metric-DP noise injection.
\vspace{-0.1in}
\begin{table*}[h]
\centering
\caption{Average F1 scores on AML datasets across high-illicit (HI) and low-illicit (LI) ratio settings. Best results in bold.} 
\begin{tiny}
\resizebox{1.0\textwidth}{!}{
\begin{tabular}{l|c|c|c|c|c|c}
\toprule 
\cmidrule(lr){2-4}
& HI-Small & HI-Medium  & HI-Large & LI-Small & LI-Medium  & LI-Large    \\
\midrule 
SC-GIN &$0.1526\pm0.0157$ & $0.3572\pm0.0305$   & $0.3459\pm0.0271$ & $0.1238\pm0.0101$ & $0.0746\pm0.0153$ & $0.0227\pm0.0132$   \\
SC-PNA & $0.4409\pm0.0294$   &  $0.5305\pm0.0311$   & $0.5744\pm0.0182$& $0.1396\pm0.0148$& $0.2472\pm0.0200$ &  $0.1262\pm0.0198$ \\
FedAvg-GIN  &   $0.4103\pm0.0335$   & $0.5421\pm0.0273$  & $0.6235\pm0.0310$  & $0.0000\pm0.0000$ & $0.0068\pm0.0012$ & $0.0000\pm0.0000$  \\ 
Swift-GIN  & $0.3873\pm0.0306$ & $0.5689\pm0.0282$&$0.6339\pm0.0205$ & $0.0000\pm0.0000$  & $0.0061\pm0.0016$ & $0.1294\pm0.0246$ \\
FedGCN-GIN & $0.4152\pm0.0291$ & $0.5538\pm0.0310$&$0.5817\pm0.0338$& $0.1702\pm0.0316$ &$0.0647\pm0.0083$&$0.0000\pm0.0000$\\
FedGNN-GIN-ST & $0.2940\pm0.0253$ & $0.5612\pm0.0297$ & $0.5210\pm0.0312$ & $0.0000\pm0.0000$ & $0.0705\pm0.0038$ & $0.0000\pm0.0000$ \\    
FedAvg-PNA   & $0.5427\pm0.0288$ & $0.5037\pm0.0341$   & $0.5958\pm0.0299$ & $0.1556\pm0.0131$ & $0.2614\pm0.0225$ & $0.0000\pm0.0000$  \\ 
Swift-PNA & $0.5746\pm0.0302$ & $0.5722\pm0.0258$ &$0.6144\pm0.0372$& $0.1995\pm0.0206$ &$0.2001\pm0.0280$ & $0.0000\pm0.0000$\\
FedGCN-PNA & $0.5653\pm0.0365$ & $0.6292\pm0.0283$ & $0.6028\pm0.0324$& $0.1726\pm0.0239$&$0.1821\pm0.0197$&$0.0000\pm0.0000$\\
FedGNN-PNA-ST & $0.5436\pm0.0311$ & $0.5382\pm0.0276$ & $0.4844\pm0.0465$ & $0.1614\pm0.0142$ & $0.1129\pm0.0173$ &$0.0000\pm0.0000$  \\
\hline 
CE-FedGNN-GIN    & $0.4916\pm0.0218$   &$0.6024\pm0.0314$  & $0.6461\pm0.0306$ &$0.1630\pm0.0155$  & $0.0828\pm0.0127$ & $0.1917\pm0.0147$   \\
CE-FedGNN-PNA   &  $\mathbf{0.6623\pm0.0273} $ & $\mathbf{0.6517\pm0.0322}$ & $\mathbf{0.7114\pm0.0251}$  &$\mathbf{0.2655\pm0.0199}$ & $\mathbf{0.2918\pm0.0106}$ & $\mathbf{0.3158\pm 0.0215}$  \\
\bottomrule 
\end{tabular} 
}
\end{tiny} 
\label{tab:results_aml}
\end{table*}

\begin{table*}[htbp] 
\centering
\caption{Average F1 scores on citation network benchmarks. Best results in bold.}
\begin{tiny} 
\resizebox{0.8\textwidth}{!}{
\begin{tabular}{l|c|c|c|c}
\toprule 
\cmidrule(lr){2-5}
  & Cora & Citeseer  & Pubmed & MSAcademic  \\
\midrule 
SC& $0.2511\pm0.0012$ & $0.2049\pm0.0015$ & $0.3792\pm0.0020$ & $0.7614\pm0.0044$\\
FedAvg& $0.1868\pm0.0009$ & $0.2189\pm0.0014$ & $0.2855\pm0.0029$ & $0.2759\pm0.0023$\\
FedSage+& $0.1876\pm0.0078$ & $0.1128\pm0.0006$ & $0.2705\pm0.0005$ & $0.1287\pm0.0013$\\ 
FedPUB
 & $0.2782\pm0.0000$ & $0.3696\pm0.0000$ & $0.3953\pm0.0000$ & $0.8462\pm0.0000$  \\ 
Swift&$0.4296\pm0.0016$ &$0.4023\pm0.0018$ & $0.6181\pm0.0012$ & $0.8003\pm0.0015$\\
FedGCN&$0.4005\pm0.0024$ & $0.4042\pm0.0029$ & $0.3614\pm0.0025$ & $0.7935\pm0.0017$\\
FedGNN-ST & $0.4345\pm0.0019$ & $0.3972\pm0.0021$ & $0.6347\pm0.0018$  & $0.8168\pm0.0022$ \\
\hline 
CE-FedGNN    & $\mathbf{0.4701\pm0.0020}$ & $\mathbf{0.4343\pm0.0034}$ & $\mathbf{0.6517\pm0.0023}$ & $\mathbf{0.8497\pm0.0028}$ \\
\bottomrule 
\end{tabular} 
} 
\end{tiny} 
\label{tab:results_citation}
\end{table*}  
\vspace{-0.1in} 
\paragraph{Data.}
Real-world anti--money laundering (AML) transaction data are not publicly available due to regulatory constraints. We therefore adopt the realistic transaction simulator of \citet{altman2023realistic}, which generates financial networks by modeling entities such as banks and individuals and injecting illicit activity through well-established laundering patterns. This reliance on synthetic data reflects the practical motivation for federated learning in AML, where direct data sharing is inherently restricted. We additionally evaluate on four widely used citation network benchmarks: Cora~\citep{sen2008collective}, Citeseer~\citep{sen2008collective}, PubMed~\citep{namata2012query}, and MSAcademic~\citep{shchur2018pitfalls}.

For the AML task, we consider small, medium, and large datasets, each provided in two variants with high (HI) or low (LI) illicit transaction ratios (e.g., \textit{HI-Small}). Following \citet{altman2023realistic}, we use a temporal train--validation--test split. To simulate federated deployment, data are partitioned across 4--32 clients. For cross-client edges, each client retains a copy of the corresponding edge attributes, reflecting realistic inter-bank transactions. Detailed dataset statistics are provided in Appendix~\ref{app:data_statistics}.
\vspace{-0.1in}
\paragraph{Setup.}
Although the proposed method applies to a broad class of message-passing GNNs (Section~\ref{sec:statement}), we focus our AML experiments on GIN~\citep{xu2018powerful}, which achieves maximal expressive power among message-passing GNNs, and PNA~\citep{corso2020principal}, which combines multiple aggregators in parallel. Both models are augmented with edge updates, ego IDs, and port numbering following \citet{egressy2024provably}. For citation network benchmarks, we use a standard GCN. Each architecture consists of two message-passing layers. We use the average testing F1 score across clients as the evaluation metric.

We compare against the following baselines:
(1) \textbf{Single Client (SC)}, which trains a local model independently on each client;
(2) \textbf{FedAvg}~\citep{mcmahan2017communication}, which communicates only model parameters;
(3) \textbf{Swift-FedGNN}~\citep{qiuswift}, which intermittently trains on the global graph while primarily relying on local graphs;
(4) \textbf{FedGCN}~\citep{hu2022fedgcn}, which shares node embeddings only once at initialization;
(5) \textbf{FedGNN-ST} is a naive federated version of \citep{fey2021gnnautoscale}, which shares stale node embeddings at communication rounds (without moving average for the embedding estimators); 
(6) \textbf{FedSage+}~\citep{zhang2021subgraph}, which generates missing nodes from local subgraphs; and
(7) \textbf{FedPUB}, which learns weighted aggregation of model updates.
FedSage+ incurs substantial computational overhead due to on-the-fly node generation, while FedPUB requires full-batch computation. Neither method scales to large graphs, so we include them only on citation network benchmarks. Hyperparameter setups are described in Appendix~\ref{app:data_statistics}.

\paragraph{Primary results.}
Table~\ref{tab:results_aml} reports results on AML datasets. On HI datasets, our method consistently outperforms all baselines across scales, with CE-FedGNN-PNA achieving the best overall performance. All methods experience performance degradation in the LI setting due to severe class imbalance, but CE-FedGNN maintains a clear advantage where competing methods either perform substantially worse or collapse to trivial predictions. These results indicate strong robustness under extreme imbalance.

Table~\ref{tab:results_citation} reports performance on citation network benchmarks. CE-FedGNN consistently outperforms scalable federated baselines. While FedPUB performs strongly on small graphs, it relies on full-batch training and does not scale to large graphs or realistic federated settings. CE-FedGNN attains competitive performance while maintaining communication efficiency and scalability, demonstrating that the proposed method generalizes beyond financial transaction graphs. 

\vspace{-0.1in}
\begin{figure}[htbp]
    \centering
    \includegraphics[width=0.23\linewidth]{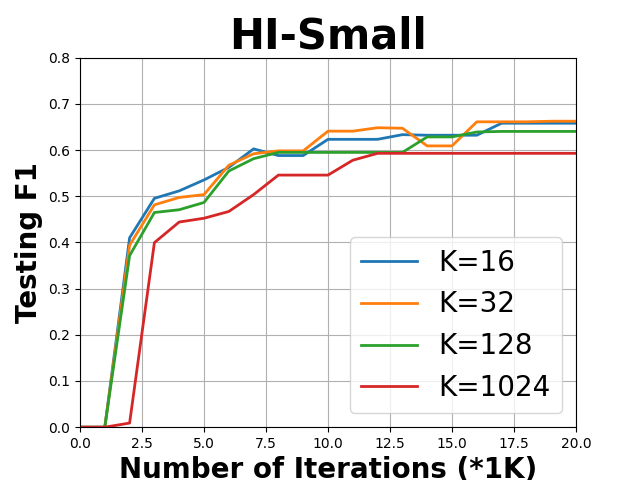}
    \includegraphics[width=0.23\linewidth]{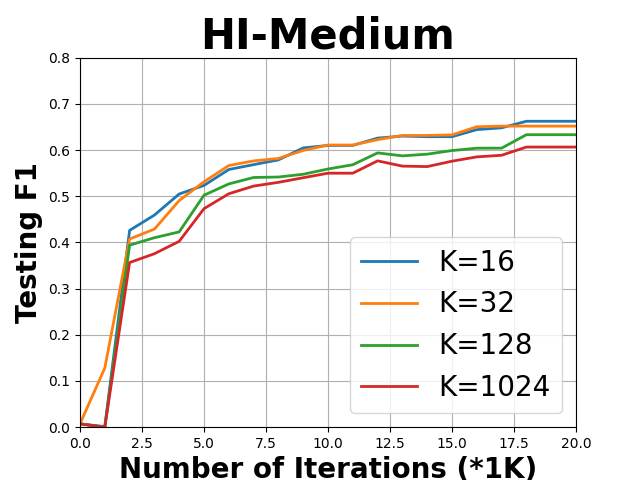}
    \includegraphics[width=0.23\linewidth]{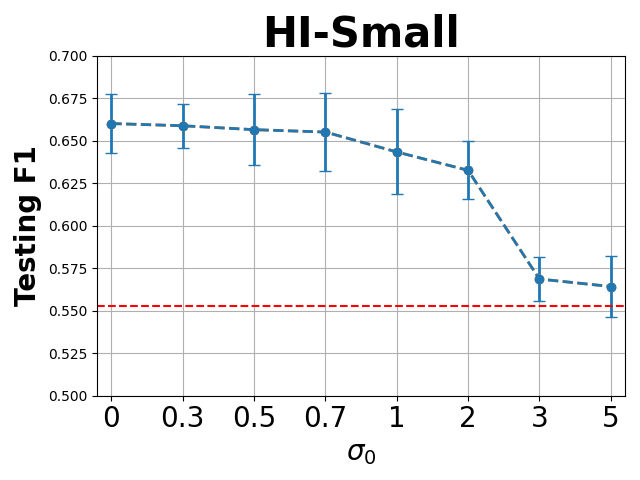}
    \includegraphics[width=0.23\linewidth]{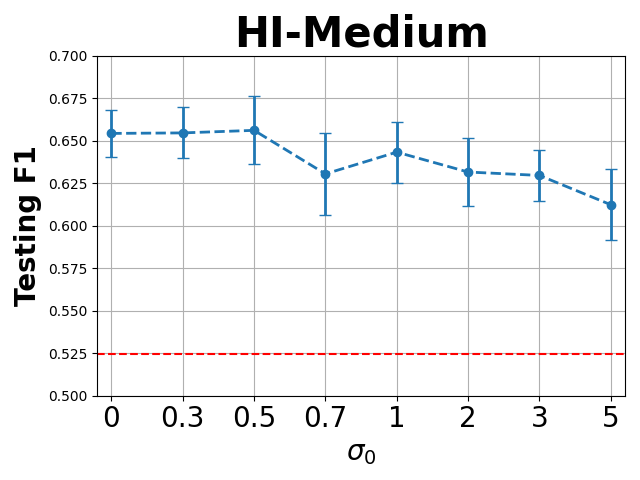} 
    \caption{Left two panels: Effect of communication interval $K$ on HI-Small and HI-Medium. Right two panels: Performance under varying embedding noise $\sigma_0$. The red dashed line indicates the FedAvg baseline.} 
    \label{fig:vark_K}
    \vspace{-0.21in}
\end{figure}

\paragraph{Communication efficiency.}
Figure~\ref{fig:vark_K} (left two panels) studies the effect of the communication interval $K$ on CE-FedGNN-PNA. Increasing $K$ substantially reduces communication frequency yet has limited impact on predictive performance. Even at $K=1024$, the model maintains strong accuracy and continues to outperform the baselines reported in Table~\ref{tab:results_aml}. This confirms that the proposed moving-average mechanism enables effective reuse of embeddings across many local steps.
\paragraph{Privacy--utility tradeoff.}
We examine the impact of Gaussian noise injection on model performance. Since the effects of noise on shared model parameters and gradients are extensively studied in standard FL, we focus on perturbations applied to shared node embeddings, the primary target of our metric-DP analysis. We fix $\sigma_1 = \sigma_2 = 10^{-3}$ for model and gradient updates and vary the embedding noise level $\sigma_0$, while clipping embedding norms to $5$ and model norms to $15$, with $K=32$ and $T=30\text{k}$.
Figure~\ref{fig:vark_K} (right two panels) shows a clear utility--noise tradeoff: moderate noise levels have a limited effect on performance, while larger $\sigma_0$ values lead to noticeable degradation. Notably, CE-FedGNN remains competitive with FedAvg even under relatively strong embedding noise, suggesting that the proposed aggregation and smoothing mechanisms are robust to noise injected. 

Additional experiments on ablation of algorithmic components, communication efficiency, performance under attribute inference attacks, as well as mapping between the noise level and metric-DP guarantee are reported in Appendix~\ref{sec:app:additional}.

\vspace{-0.1in} 
\section{Conclusion}
\vspace{-0.1in}
\label{sec:conclusion}
We studied FL on graph data with cross-client coupling. We proposed CE-FedGNN, a communication-efficient and privacy-preserving federated GNN framework that infrequently exchanges aggregated node embeddings to capture the global graph structure while substantially reducing communication overhead. We established convergence guarantees for the resulting multi-layer compositional optimization problem and provided $(\eps,\delta)$-metric-DP guarantees for released embeddings under a public-cohort threat model, yielding meaningful privacy at noise levels where standard DP is vacuous. Experiments on synthetic interbank anti--money laundering benchmarks and real-world citation networks demonstrate that CE-FedGNN achieves strong performance across dataset scales and remains robust under infrequent communication and privacy-preserving noise.
\vspace{-0.1in}
\section{Impact and Limitations}
\label{sec:impact_limitations} 
CE-FedGNN enables FL GNN training across organizations, with applications including financial fraud 
detection and cross-institutional analytics. Such systems are dual-use: 
improved detection can also enable surveillance without appropriate 
oversight, underscoring the importance of formal privacy guarantees.
Subsampling amplification could be further exploited to enhance privacy. AML experiments use synthetic data~\citep{altman2023realistic} 
due to regulatory restrictions on real transaction data.

\bibliography{ref,graph,privacy}

\clearpage
\appendix

\section{Graph Neural Networks}
\label{appx:GNN}

We summarize the formulations of several commonly used message-passing GNN architectures. Let $\mathcal{N}(v)$ denote the neighbor set of node $v$, and let $\phi(\cdot)$ be a nonlinear activation function.

\begin{equation}
\small
\begin{split}
& \text{GCN:}\quad
\mathbf{h}_v^{(l)} = \phi\!\Bigg( \sum_{u \in \mathcal{N}(v)\cup\{v\}}
\frac{1}{\sqrt{\tilde d_v \tilde d_u}}\,\mathbf{W}^{(l)} \mathbf{h}_u^{(l-1)} \Bigg),
\quad \tilde A = A+I,\ \tilde d_v = \sum_{u} \tilde A_{vu}. \\
& \text{GIN:}\quad
\mathbf{h}_v^{(l)} = \phi\!\left( \mathbf{W}^{(l)} \Big( (1+\varepsilon)\mathbf{h}_v^{(l-1)}
+ \sum_{u \in \mathcal{N}(v)} \mathbf{h}_u^{(l-1)} \Big) \right),
\quad \varepsilon^{(l)}\ \text{is fixed or learnable}. \\
& \text{GraphSAGE-mean:}\quad
\mathbf{h}_v^{(l)} = \phi\!\Bigg( \mathbf{W}^{(l)} \cdot
\frac{1}{|\mathcal{N}(v)\cup\{v\}|} \sum_{u \in \mathcal{N}(v)\cup\{v\}} \mathbf{h}_u^{(l-1)} \Bigg).
\end{split}
\end{equation} 

We denote $h^{(l)}_v = \phi(\th^{(l)}_v) = \phi(\W^{(l)} \hh^{(l)}_v)$, consistent with the notation used in the main text.

Recent advances in GNN research have studied how aggregation functions~\citep{xu2018powerful} and spectral perspectives~\citep{wang2022powerful} affect expressive power, enabling models to capture higher-order dependencies and subtle structural patterns. To further enhance expressivity, several architectural extensions have been proposed, including port numbering, ego identifiers, and reverse message passing~\citep{jaume2019edgnn,sato2019approximation,you2021identity}.

These techniques are particularly relevant for transaction networks, which often take the form of directed multigraphs with repeated interactions between entities~\citep{cardoso2022laundrograph,kanezashi2022ethereum,weber2018scalable,weber2019anti,nicholls2021financial}. Building on these ideas, \citet{egressy2024provably} demonstrate both theoretically and empirically that expressive GNN architectures substantially improve the detection of money laundering patterns in transaction graphs. In this work, we focus on enabling such expressive GNN models to be trained in a federated setting, where graph data are distributed across clients and subject to privacy constraints.

\section{Convergence Analysis}
\label{app:convergence}

A function $f$ is said to be $C_0$-Lipschitz continuous if $\|f(\x) - f(\x')\| \leq C_0\|\x - \x'\|$ for all $\x, \x'$, and $C_1$-smooth if $\|\nabla f(\x) - \nabla f(\x')\| \leq C_1\|\x - \x'\|$. The superscripts $H$ and $h$ denote forward and backward propagation computed using the embedding estimator $H$ and the true embeddings of sampled data, respectively.

\subsection{Proof of Lemma~\ref{lem:H_estimate}}

Let $p_i(v)$ denote the probability that a node $v$ is sampled on client $i$. For edge-based prediction tasks, this satisfies $p_i(v) \geq B_0 / |\mathcal{E}_i|$, where $B_0$ is the number of seed edges sampled per iteration and $|\mathcal{E}_i|$ is the total number of edges on client $i$. Let $\th^{(l)}_{r,k}(v)$ denote the deterministic pre-activation embedding at layer $l$, computed using the averaged model $\W_{r,k} := \frac{1}{N}\sum_i \W_{r,k,i}$ and the full neighborhood of $v$. We use $c(v)$ to denote the hosting client of node $v$.

For $l > 1$, by the update rule of the moving-average estimator,
\begin{small}
\begin{small}
\begin{equation}
\begin{split}
&\E\|\tH^{(l)}_{r,k,i}(v) - \th^{(l)}_{r,k}(v)\|^2 \\   
&= p(v) \E \bigg\| (1-\gamma) \tH^{(l)}_{r,k-1,i}(v) 
+ \gamma \frac{1}{n_{r,k,i}(v)}\Bigg(\W^{(l)}_{r,k-1,i}\cdot \sum\limits_{u\in \N_{r,k,i}(v)} H^{(l-1)}_{r,k-1,i}(u)  \Bigg) \\ 
&~~~~~~ - \th^{(l)}_{r,k-1}(v) + \th^{(l)}_{r,k-1}(v) - \th^{(l)}_{r,k}(v)\bigg\|^2 
+ (1-p(v)) \E \left\| \tH^{(l)}_{r,k-1,i}(v) -  \th^{(l)}_{i,k}(v) \right\|^2 \\  
& = p(v) \E \bigg\| (1-\beta) (\tH^{(l)}_{r,k-1,i}(v) - \th^{(l)}_{r,k-1}(v)) \\
&~~~ + \gamma  \Bigg( \W^{(l)}_{r,k-1,i}\cdot\frac{1}{n_{r,k,i}}\Bigg(\sum\limits_{u\in \N_{r,k,i}(v)} H^{(l-1)}_{r,k-1,i}(u) \Bigg) - \th^{(l)}_{r,k-1}(v) \Bigg) \\
&~~~~~~~~~+ \th^{(l)}_{r,k-1}(v) - \th^{(l)}_{r,k}(v)\bigg\|^2 
+ (1-p(v)) \E \left\| \tH^{(l)}_{r,k-1,i}(v) -  \th^{(l)}_{i,k}(v) \right\|^2 \\ 
&\leq (1+\frac{\gamma}{4})p(v) \E \bigg\| (1-\beta) (\tH^{(l)}_{r,k-1,i}(v) - \th^{(l)}_{r,k-1}(v)) \\
&~~~ + \gamma \Bigg(\W^{(l)}_{r,k-1,i}\cdot \frac{1}{n_{r,k,i}(v)}\Bigg(\sum\limits_{u\in \N_{r,k,i}(v)} H^{(l-1)}_{r,k-1,i}(u)\Bigg) - \th^{(l)}_{r,k-1}(v) \Bigg)\bigg\|^2 \\
&~~~+ (2+\frac{4}{\gamma}+\frac{4}{\gamma p(v)})p(v) \bigg\|\th^{(l)}_{r,k-1}(v) - \th^{(l)}_{r,k}(v)\bigg\|^2 
+ (1-p(v)) (1+\frac{\gamma p(v)}{4}) \E \left\| \tH^{(l)}_{r,k-1,i}(v) -  \th^{(l)}_{i,k-1}(v) \right\|^2 \\ 
\end{split}    
\end{equation}  
which then leads to 
\begin{equation}
\begin{split}
& \E\|\tH^{(l)}_{r,k,i}(v) - \th^{(l)}_{r,k}(v)\|^2 \leq  (1+\frac{\gamma}{4})p(v) \E \bigg\| (1-\gamma) (\tH^{(l)}_{r,k-1,i}(v) - \th^{(l)}_{r,k-1}(v)) \\
&~~~ + \gamma \Bigg( \W^{(l)}_{r,k-1,i}\cdot \bigg( \frac{1}{n_{r,k,i}(v)}\sum\limits_{u\in \N_{r,k,i}(v)} H^{(l-1)}_{r,k-1,i}(u)  - \frac{1}{n_{i}(v)} \sum\limits_{u\in \N_i(v)} H^{(l-1)}_{r,k-1,i}(u)  \bigg) \Bigg)\\ 
&~~~ + \gamma\Bigg(\W^{(l)}_{r,k-1,i}\cdot\bigg(\frac{1}{n_{i}(v)} \sum\limits_{u\in \N_i(v)} H^{(l-1)}_{r,c(u),k-1}(u)  - \th^{(l)}_{r,k-1}(v)\bigg) \Bigg)\bigg\|^2 \\ 
&~~~+ \frac{5}{\gamma} \bigg\|\th^{(l)}_{r,k-1}(v) - \th^{(l)}_{r,k}(v)\bigg\|^2 
+ (1-p(v)) (1+\frac{\gamma p(v)}{4}) \E \left\| \tH^{(l)}_{r,k-1,i}(v) -  \th^{(l)}_{i,k-1}(v) \right\|^2. 
\end{split}    
\end{equation}  
\end{small} 
Then 
noting 
\begin{equation}
\begin{split}
\E_{r,k-1}\bigg( \frac{1}{n_{r,k,i}(v)}\sum\limits_{u\in \N_{r,k,i}(v)} H^{(l-1)}_{r,k-1,i}(u)  - \frac{1}{n_{i}(v)} \sum\limits_{u\in \N_i(v)} H^{(l-1)}_{r,c(u),k-1}(u)  \bigg) = 0, 
\end{split}
\end{equation}
we have 
\begin{equation}
\begin{split}
&\E\|\tH^{(l)}_{r,k,i}(v) - \th^{(l)}_{r,k}(v)\|^2 \\ 
&\leq (1+\frac{\gamma}{4})p(v) \E \bigg\| (1-\gamma) (\tH^{(l)}_{r,k-1,i}(v) - \th^{(l)}_{r,k-1}(v)) \\
&~~~ + \gamma\Bigg(\W^{(l)}_{r,k-1,i} \cdot  \frac{1}{n_{i}(v)} \sum\limits_{u\in \N(v)} H^{(l-1)}_{r,k-1,i}(u)  - \th^{(l)}_{r,k-1}(v) \Bigg)\bigg\|^2 \\ 
&~~~+ \gamma^2 G^2  + \frac{5}{\gamma} \bigg\|\th^{(l)}_{r,k-1}(v) - \th^{(l)}_{r,k}(v)\bigg\|^2 
+ (1-p(v)) (1+\frac{\gamma p(v)}{4}) \E \left\| \tH^{(l)}_{r,k-1,i}(v) -  \th^{(l)}_{i,k-1}(v) \right\|^2  \\   
& \leq (1+\frac{\gamma}{4})^2(1-\gamma)^2 p(v) \E\left\| \tH^{(l)}_{r,k-1,i}(u) -  \th^{(l)}_{i,k-1}(u) \right\|^2 \\
&~~~ + (1+\frac{4}{\gamma})\gamma^2\|\Bigg( \W^{(l)}_{r,k-1,i}\cdot \frac{1}{n_{i}(v)}(\sum\limits_{u\in \N(v)} H^{(l-1)}_{r,k-1,i}(u)   - \th^{(l)}_{r,k-1}(v)) \Bigg)\|^2  + \gamma^2 G^2 \\
&~~~ + \frac{5}{\gamma} \bigg\|\th^{(l)}_{r,k-1}(v) - \th^{(l)}_{r,k}(v)\bigg\|^2 
+ (1-p(v)) (1+\frac{\gamma p(v)}{4}) \E \left\| \tH^{(l)}_{r,k-1,i}(v) -  \th^{(l)}_{i,k-1}(v) \right\|^2 \\   
&\leq (1-\frac{\gamma p(v)}{4}) \E\left\| \tH^{(l)}_{r,k-1,i}(u) -  \th^{(l)}_{i,k-1}(u) \right\|^2 
+ 8\gamma C_W^2 \frac{1}{n_i(v)} \sum\limits_{u\in \N(v)} \|\tH^{(l-1)}_{r,k-1,i}(u) - \th^{(l-1)}_{r,k-1}(u)\|^2) \\
&~~~+\gamma^2 C_W^2 C_H^2 + \frac{8}{\gamma} C^2 \|\W_{r,k} - \W_{r,k-1}\|^2. 
\end{split}    
\end{equation}  
where $C$ is a Lipschitz constant of $\tilde{h}^{(l)}_{r,k}$ over $\W$, depending on the constants in Assumption \ref{ass:1}.
\end{small} 
Rearranging the terms, we have 
\begin{equation}
\begin{split}
&\E\left\| \tH^{(l)}_{r,k-1,i}(v) -  \th^{(l)}_{i,k-1}(v) \right\|^2 \leq \frac{4\left(\E\left\| \tH^{(l)}_{r,k-1,i}(v) -  \th^{(l)}_{i,k-1}(v) \right\|^2 - \E\left\| \tH^{(l)}_{r,k,i}(v) -  \th^{(l)}_{i,k}(v) \right\|^2\right)}{\gamma p(v)} \\
& +  \frac{32 C_W^2}{p(v) n_i(v)} \sum\limits_{u\in \N(v)} \|\tH^{(l-1)}_{r,k-1,i}(u) - \th^{(l-1)}_{r,k-1}(u)\|^2) +\frac{4}{p(v)}\gamma C_W^2 C_H^2 + \frac{32}{\gamma^2 p(v)} C^2 \|\W_{r,k} - \W_{r,k-1}\|^2. 
\end{split}
\end{equation}

Note that $\tH^{(l)}_{r,i,0} = \tH^{(l)}_{r-1,i,K}$, and $\th^{(l)}_{r,0} = \th^{(l)}_{r-1,K}$. Let $p:=\min\limits_{u\in\V} p(u)$. 
Taking the telescoping sum, we have 
\begin{equation}
\begin{split}
&\frac{1}{RK}\sum_{r}\sum_{k}\E\left\| \tH^{(l)}_{r,k-1,i}(v) -  \th^{(l)}_{i,k-1}(v) \right\|^2 \leq O\Bigg(\frac{4(L+1)C^2}{\gamma p RK} \\ 
&~~~+\frac{4}{p}\gamma (L+1) G^2  + \frac{32}{\gamma^2 p} (L+1) C^2 \|\W_{r,k} - \W_{r,k-1}\|^2\Bigg). 
\end{split}
\end{equation} 

And thus using Lipschitz of $\phi(\cdot)$, we obtain
\begin{equation}
\begin{split}
&\frac{1}{RK}\sum_{r}\sum_{k}\E\left\| H^{(l)}_{r,k-1,i}(v) -  h^{(l)}_{i,k-1}(v) \right\|^2 \leq O\Bigg(\frac{4(L+1)C^2}{\gamma p RK} \\ 
&~~~+\frac{4}{p}\gamma (L+1) G^2 + \frac{32}{\gamma^2 p} (L+1) C^2 \|\W_{r,k} - \W_{r,k-1}\|^2\Bigg). 
\end{split}
\end{equation} 

\subsection{Proof of Lemma~\ref{lem:G_estimate}}
By the update rule of the gradient estimator~\eqref{eq:G_ma},
\begin{equation}
\begin{split}
&\|\bar{G}_{r,k} - \nabla F(\W_{r,k}) \|^2 \\
& \leq (1+\frac{\beta}{4})\bigg\| (1-\beta) \bar{G}_{r,k-1} + \beta\frac{1}{N} \sum\limits_{i=1}^{N} \left( \hat{\nabla} F_i(\W_{r,k-1,i};B_{r,k,i})\right) - \nabla F(\W_{r,k-1})  \bigg\|^2 \\
&~~~ + (1+\frac{4}{\beta}) C^2 \|\W_{r,k-1} - \W_{r,k}\|^2  \\  
& =  (1+\frac{\beta}{4}) \bigg\| (1-\beta) (\bar{G}_{r,k-1} - \nabla F(\W_{r,k-1})) \\
&~~~ + \beta \frac{1}{N} \sum\limits_{i=1}^{N} \bigg( \hat{\nabla} F_i^H(\W_{r,k-1,i};B_{r,k,i})  -  \hat{\nabla} F_i^h(\W_{r,k-1,i};B_{r,k,i}) \bigg) \\
&~~ + \beta \frac{1}{N} \sum\limits_{i=1}^{N} \bigg(  \hat{\nabla} F_i^h(\W_{r,k-1,i};B_{r,k,i}) -  \nabla F(\W_{r,k-1}) \bigg ) \bigg\|^2 + (1+\frac{4}{\beta}) C^2 \|\W_{r,k-1} - \W_{r,k}\|^2  \\ 
& = (1+\frac{\beta}{4}) \bigg\| (1-\beta) (\bar{G}_{r,k-1} - \nabla F(\W_{r,k-1})) \\
&~~ + \beta  \frac{1}{N} \sum\limits_{i=1}^{N} \bigg( \hat{\nabla} F_i^h(\W_{r,k-1,i};B_{r,k,i})  -  \nabla F(\W_{r,k-1})  \bigg) \bigg\|^2 \\
&~~~ + (1+\frac{4}{\beta}) \beta^2\frac{1}{N} \sum\limits_{i=1}^{N}\bigg\| \bigg( \hat{\nabla} F_i^H(\W_{r,k-1,i};B_{r,k,i})  -  \hat{\nabla} F_i^h(\W_{r,k-1,i};B_{r,k,i}) \bigg) \|^2 \\
&~~~ + (1+\frac{4}{\beta}) C^2 \|\W_{r,k-1} - \W_{r,k}\|^2   \\
& \leq (1+\frac{\beta}{4})^2 \bigg\| (1-\beta) (\bar{G}_{r,k-1} - \nabla F(\W_{r,k-1})) \\
&~~ + \beta  \frac{1}{N} \sum\limits_{i=1}^{N} \bigg( \hat{\nabla} F_i^h(\W_{r,k-1};B_{r,k,i})  -  \nabla F(\W_{r,k-1})  \bigg) \bigg\|^2 + (1+\frac{4}{\beta}) \beta^2\frac{1}{N} \sum\limits_{i=1}^{N} \|\W_{r,k-1}-\W_{r,k-1,i}\|^2 \\
&~~~ + (1+\frac{4}{\beta}) \beta^2\frac{1}{N} \sum\limits_{i=1}^{N}\bigg\| \bigg( \hat{\nabla} F_i^H(\W_{r,k-1,i};B_{r,k,i})  -  \hat{\nabla} F_i^h(\W_{r,k-1,i};B_{r,k,i}) \bigg) \|^2 \\
&~~~ + (1+\frac{4}{\beta}) C^2 \|\W_{r,k-1} - \W_{r,k}\|^2   \\
&\leq (1-\frac{\beta}{4}) \|\bar{G}_{r,k-1} - \nabla F(\W_{r,k-1})\|^2  + \frac{\beta^2 \sigma^2}{N} + C^2 5\beta \eta^2 K^2 D^2  \\
&~~~ + 5\beta \frac{1}{N} \sum\limits_{i=1}^{N} \| \hat{\nabla} F_i^{H}(\W_{r,k-1,i};B_{r,k,i})  -  \hat{\nabla} F_i^h(\W_{r,k-1,i};B_{r,k,i}) \|^2 + \beta^3 K^2 C_H^2 \\
&~~~ + \frac{5C^2 \eta^2}{\beta} \|\bar{G}_{r,k-1}\|^2,
\end{split} 
\end{equation}  
where superscript $H'$ denotes plugging all $H$ estimators at iteration $(r,i,k)$. For remote nodes, the difference between current estimator and previous estimator is bounded by $\beta^2 K^2 C_H^2$. Moreover,
\begin{equation}
\begin{split}
\| \hat{\nabla} F_i^{H'}(\W_{r,k-1,i};B_{r,k,i})  -  \hat{\nabla} F_i^h(\W_{r,k-1,i};B_{r,k,i}) \|^2 \leq \sum\limits_{u\in B_{r,k,i}} C^2 m(u,l) \|\tH^{(l)_{r,k-1,i}}(u) - \th^{(l)}_{r,k-1,i}(u)\|^2,
\end{split}
\label{eq:F_to_H}
\end{equation}
with $m(u,l)$ denotes the weight for node $u$ at layer $l$ satisfying $\sum\limits_{u\in B_{r,k,i}} m(u,l) = 1$. 

Rearranging terms, applying telescoping sum and substituting Lemma \ref{lem:dp_G_estimate}, 
\begin{equation}  
\begin{split} 
&\frac{1}{RK}\sum_{r}\sum_{k}\|(\bar{G}_{r,k-1} - \nabla F(\W_{r,k-1})) \|^2 \leq O\Bigg(\frac{1}{\beta RK} + \frac{1}{\gamma p RK} \\
&+ \frac{\beta}{N} + \frac{\gamma}{p} + \beta^2 K^2 + \gamma^2 K^2  + \frac{8C^2 \eta^2}{\beta^2} \|\bar{G}_{r,k-1}\|^2\Bigg).  
\end{split}  
\end{equation}  
With $\eta=O(\beta)$, we have
\begin{equation}
\begin{split}
&\frac{1}{RK}\sum_{r}\sum_{k}\|(\bar{G}_{r,k-1} - \nabla F(\W_{r,k-1})) \|^2 \leq O\Bigg(\frac{1}{\beta RK} + \frac{1}{\gamma pRK} \\
&+ \frac{\beta}{N} + \frac{\gamma}{p} + \beta^2 K^2 + \frac{\gamma^2 K^2}{p} + \frac{\eta^2}{\beta^2} \|\nabla F(\W_{r,k-1})\|^2\Bigg).  
\end{split}  
\end{equation}

\subsection{Proof of Theorem \ref{thm}}
Using $C_1$-smoothness of $F$, we have
\begin{equation}
\begin{split}
F(\W_{r}) &\leq F(\W_{r-1}) + \langle \nabla F(\W_{r-1}), \W_r-\W_{r-1}\rangle + \frac{C_1}{2} \|\W_{r} - \W_{r-1}\|^2 \\
& = F(\W_{r-1}) - \langle \nabla F(\W_{r-1}), \eta \frac{1}{N} \sum_i \sum_k G_{r,k,i}(\W_{r,k,i})  \rangle + \frac{C_1}{2} \|\W_{r} - \W_{r-1}\|^2 \\
& = F(\W_{r-1}) - \langle \nabla F(\W_{r-1}), \eta \frac{1}{N} \sum_i \sum_k \hat{\nabla} F_i(\W_{r-1}) \rangle + \frac{C_1}{2} \|\W_{r} - \W_{r-1}\|^2  \\  
&~~~ - \langle \nabla F(\W_{r-1}), \eta \frac{1}{N} \sum_i \sum_k G_{r,k,i} -   \eta \frac{1}{N} \sum_i \sum_k \hat{\nabla} F_i(\W_{r-1}) \rangle  \\
& = F(\W_{r-1}) - \eta K \|F(\W_{r-1})\|^2 + \frac{\eta}{2} K \|\nabla F(\W_{r-1})\|^2
\\
&~~~ + \frac{\eta}{K} \left\| \frac{1}{N} \sum_i \sum_k G_{r,k,i} - \frac{1}{N} \sum_i \sum_k \hat{\nabla} F_i(\W_{r-1})  \right\|^2 \\
&~~~ + \frac{\eta K}{NK}\sum_i \sum_k \|\W_{r-1} - \W_{r,k,i}\|^2 
+ \frac{C_1}{2} \eta^2 \left\| \frac{1}{N} \sum_i \sum_k G_{r,k,i} \right\|^2.   
\end{split}    
\end{equation} 

Thus, plugging Lemma \ref{lem:G_estimate}, we have
\begin{equation}
\begin{split}
&\frac{1}{R}\sum\limits_{r} \E\|\nabla F(\W_{r-1})\|^2 \\
& \leq O\Bigg(\frac{F(\W_0) - F(\W_*)}{\eta R K} +\frac{1}{RK}\sum_{r}\sum_{k}\|(\bar{G}_{r,k-1} - \nabla F(\W_{r,k-1})) \|^2
+\eta^2 K^2 D^2 \Bigg)\\
& \leq O(\frac{1}{\eta RK} + \frac{1}{\beta RK} +\frac{1}{\gamma pRK} + \frac{\beta}{N} + \frac{\gamma}{p} + \beta^2K^2).  
\end{split}
\end{equation}

\section{Privacy Analysis}
\label{app:privacy}

This appendix contains (i) proofs of the noise-perturbed convergence lemmas and theorem stated in Section~\ref{sec:privacy}, and (ii) the proof of the metric-DP composition theorem (Theorem~\ref{thm:mdp_composition}).

\subsection{Convergence under noise perturbation}

\begin{lemma}
\label{lem:dp_H_estimate}
Under Assumption~\ref{ass:1}, the noise-perturbed version of Algorithm~\ref{alg:ce_fedgnn} ensures
\begin{equation*}
\small
\E\!\left[\| H^{(l)}_{r,k,i}(u) - h^{(l)}(u) \|^2\right]
\le O\!\bigg(\frac{1}{p \gamma R K}+ \gamma+ \beta^2 K^2
+ \frac{\|\mathbf{W}^{(l)}_{r,k} - \mathbf{W}^{(l)}_{r,k-1}\|^2}{\gamma} + \sigma_0^2 + \frac{\sigma_1}{\gamma}\bigg).
\end{equation*}
\end{lemma}

\begin{lemma}
\label{lem:dp_G_estimate}
Under Assumption~\ref{ass:1}, the noise-perturbed version of Algorithm~\ref{alg:ce_fedgnn} ensures
\begin{equation*}
\small
\E\!\left[\left\|\frac{1}{N} \sum_{i=1}^N G_{r,k,i} - \nabla F(\mathbf{W}_{r,k})\right\|^2\right]
\le O\!\bigg(\frac{1}{\gamma R K} + \gamma + \beta^2 K^2 + \frac{\|\mathbf{W}^{(l)}_{r,k} - \mathbf{W}^{(l)}_{r,k-1}\|^2}{\gamma} + \sigma_0^2 + \frac{\sigma_1 + \sigma_2}{\beta}\bigg).
\end{equation*}
\end{lemma}

The proof of Lemma~\ref{lem:dp_H_estimate} mirrors that of Lemma~\ref{lem:H_estimate}. Within a round $r$, the update rule with embedding noise gives
\begin{equation}
\small
\begin{split}
&\E\|\tH^{(l)}_{r,k,i}(v) - \th^{(l)}_{r,k}(v)\|^2 \\   
&= p(v) \E \bigg\| (1-\gamma) \tH^{(l)}_{r,k-1,i}(v) 
+ \gamma \frac{1}{n_{r,k,i}(v)}\Bigg(\W^{(l)}_{r,k-1,i}\cdot \sum\limits_{u\in \N_{r,k,i}(v)} H^{(l-1)}_{r,k-1,i}(u)  \Bigg) + \gamma \N(0,\sigma_0^2 I)\\ 
&~~~~~~ - \th^{(l)}_{r,k-1}(v) + \th^{(l)}_{r,k-1}(v) - \th^{(l)}_{r,k}(v)\bigg\|^2 
+ (1-p(v)) \E \left\| \tH^{(l)}_{r,k-1,i}(v) -  \th^{(l)}_{i,k}(v) \right\|^2 \\  
&\leq (1+\frac{\gamma p(v)}{8}) (1+\frac{\gamma}{4})p(v) \E \bigg\| (1-\beta) (\tH^{(l)}_{r,k-1,i}(v) - \th^{(l)}_{r,k-1}(v)) \\
&~~~ + \gamma \Bigg(\W^{(l)}_{r,k-1,i}\cdot \frac{1}{n_{r,k,i}(v)}\Bigg(\sum\limits_{u\in \N_{r,k,i}(v)} H^{(l-1)}_{r,k-1,i}(u)\Bigg) - \th^{(l)}_{r,k-1}(v) \Bigg)\bigg\|^2 + \frac{9\gamma}{p(v)} \sigma_0^2 \\
&~~~+ (2+\frac{4}{\gamma}+\frac{4}{\gamma p(v)})p(v) \bigg\|\th^{(l)}_{r,k-1}(v) - \th^{(l)}_{r,k}(v)\bigg\|^2 
+ (1-p(v)) (1+\frac{\gamma p(v)}{4}) \E \left\| \tH^{(l)}_{r,k-1,i}(v) -  \th^{(l)}_{i,k-1}(v) \right\|^2 \\ 
\end{split}    
\end{equation} 
Thus, 
\begin{equation}
\begin{split}
&\E\left\| \tH^{(l)}_{r,k-1,i}(v) -  \th^{(l)}_{i,k-1}(v) \right\|^2 \leq \frac{8\left(\E\left\| \tH^{(l)}_{r,k-1,i}(v) -  \th^{(l)}_{i,k-1}(v) \right\|^2 - \E\left\| \tH^{(l)}_{r,k,i}(v) -  \th^{(l)}_{i,k}(v) \right\|^2\right)}{\gamma p(v)} \\
& +  \frac{32 C_W^2}{p(v) n_i(v)} \sum\limits_{u\in \N(v)} \|\tH^{(l-1)}_{r,k-1,i}(u) - \th^{(l-1)}_{r,k-1}(u)\|^2) 
+  \frac{\sigma_0^2}{p(v)^2} 
+\frac{4}{p(v)}\gamma C_W^2 C_H^2 + \frac{32}{\gamma^2 p(v)} C^2 \|\W^{(l)}_{r,k} - \W^{(l)}_{r,k-1}\|^2.   
\end{split}
\end{equation} 

Because Gaussian noise is added to $\W$ at each global communication round,
we have
\begin{equation}
\begin{split}
&\E\|\tH^{(l)}_{r,K,i}(v) -  \th^{(l)}_{r,K,i}(v)\|^2 \\
&=\E\|\tH^{(l)}_{r,K,i}(v) - \th^{(l)}_{r+1,0,i}(v) +  \th^{(l)}_{r+1,0,i}(v)  -  \th^{(l)}_{i,K}(v)  \|^2 \\
&\leq \E\|\tH^{(l)}_{r,K,i}(v) - \th^{(l)}_{r+1,0,i}(v)\|^2 + C_H \sigma_1 + \sigma_1^2. 
\end{split}    
\end{equation} 
As a result, taking the telescoping sum yields 
\begin{equation}
\begin{split}
&\frac{1}{RK}\sum_{r}\sum_{k}\E\left\| \tH^{(l)}_{r,k-1,i}(v) -  \th^{(l)}_{i,k-1}(v) \right\|^2 \\
& \leq O(\frac{1}{\gamma R K} + \gamma + \gamma^2 K^2 + \frac{\|\W_{r,k} - \W_{r,k-1}\|^2}{\gamma^2} + \sigma_0^2 + \frac{\sigma_1+\sigma_1^2}{\gamma}), 
\end{split}
\end{equation}  
which concludes Lemma \ref{lem:dp_H_estimate}.

The proof of Lemma~\ref{lem:dp_G_estimate} similarly extends Lemma~\ref{lem:G_estimate}. The embedding noise contributes an additional $\sigma_0^2$ term in~\eqref{eq:F_to_H}, and the gradient noise $\mathcal{N}(0,\sigma_2^2 I)$ contributes a $\sigma_2^2 + \sigma_2/\beta$ term. Combining,
\begin{equation}
\begin{split}
&\frac{1}{RK}\sum_{r,k}\|\bar{G}_{r,k-1} - \nabla F(\W_{r,k-1}) \|^2 \\
&\leq O\!\bigg(\frac{1}{\beta RK} + \frac{1}{\gamma RK} + \beta + \gamma + \beta^2 K^2 + \frac{\eta^2}{\beta^2} \|\nabla F(\W_{r,k-1})\|^2 + \sigma_0^2 + \frac{\sigma_1+\sigma_2}{\beta}\bigg).
\end{split}
\end{equation}
Theorem~\ref{thm:dp} follows by combining this with the smoothness inequality used in the proof of Theorem~\ref{thm}.

\subsection{Proof of Theorem~\ref{thm:mdp_composition}} 

The Rényi divergence of order $\alpha$ between two spherical Gaussians $\mathcal{N}(\mu, \sigma_0^2 I)$ and $\mathcal{N}(\mu', \sigma_0^2 I)$ with equal covariance is exactly $\alpha \|\mu - \mu'\|_2^2 / (2\sigma_0^2)$~\citep{vanerven2014renyi, mironov2017renyi}. For two embeddings at $L_2$ distance $\rho$, the per-round Rényi divergence at order $\alpha$ is therefore $\alpha \rho^2 / (2\sigma_0^2)$, giving per-round $(\alpha, \alpha \rho^2/(2\sigma_0^2))$-RDP at sensitivity $\rho$.

RDP composes additively over $R'$ rounds~\citep[Prop.~1]{mironov2017renyi}, applied pointwise to each pair $(x, x')$ with $\|h_x - h_{x'}\|_2 \leq \rho$. The composed RDP converts to $(\eps, \delta)$-DP via the conversion of Canonne, Kamath, and Steinke~\citep[Prop.~12]{canonne2020discrete}, yielding
\[
\eps(\rho) = \frac{R' \alpha \rho^2}{2\sigma_0^2} + \log\frac{\alpha-1}{\alpha} - \frac{\log(\delta\alpha)}{\alpha-1}.
\]
Minimizing over $\alpha > 1$ gives the stated form. Since this bound holds uniformly for all pairs at distance at most $\rho$ in $L_2$ on the embedding space, the mechanism satisfies $(\eps(\rho), \delta)$-metric-DP at sensitivity $\rho$. 

\begin{remark}
Theorem~\ref{thm:mdp_composition} omits the subsampling amplification factor that appears in DP-SGD analyses~\citep{abadi2016deep,mironov2019r}. This is intentional: in our federated GNN protocol, the per-round update set is publicly observable (clients explicitly broadcast updated boundary-node embeddings), eliminating the adversary's uncertainty about which records contributed to a release. Settings where amplification would apply---e.g., secure aggregation of cohort updates~\citep{bonawitz2017practical} or anonymous shuffling~\citep{erlingsson2019amplification}---are complementary to our analysis and could yield additional privacy gains.
\end{remark}

\textbf{Implementation.} Theorem~\ref{thm:mdp_composition} is implemented via the Rényi DP accountant of \citet{mironov2017renyi}, available in the Opacus library~\citep{yousefpour2021opacus}, by passing noise multiplier $z = \sigma_0/\rho$ and sample rate $1$ (no subsampling amplification). The accountant is sensitivity-agnostic; only the interpretation of the returned $\eps$ as a metric-DP guarantee at distance $\rho$ requires Theorem~\ref{thm:mdp_composition}.

\section{Experimental Details}
\label{app:data_statistics} 

Dataset statistics are summarized in Table~\ref{tab:data_stats}. The ratio L/S denotes the number of edges on the largest client divided by that on the smallest client, reflecting the degree of data imbalance across clients.
\begin{table*}[tbph] 
	\caption{Statistics of the Datasets 
}\label{tab:data_stats} 
\vspace{0.1in}
	\centering 
	\scalebox{1}{\begin{tabular}{l|c|c|c|c} 
	\toprule 
	& \# of Edges & \# of Nodes & \# of Clients & L/S\\
	\hline
	HI-Small    &5,078,345  & 515,088 & 4 & 1.37 \\
	\hline 
    HI-Medium   & 31,898,238& 2,077,023 & 6 & 1.52 \\
	\hline 
	HI-Large   &179,702,229 & 2,116,168 & 32 & 3.47\\
	\hline
	LI-Small  &6,924,049 & 705,907  & 4 & 1.28\\
    \hline
	LI-Medium &31,251,483  & 2,032,095 & 6 &1.55 \\
    \hline
	LI-Large  &179,702,229 & 2,070,980  & 6 & 3.15\\ 
    \hline
    Cora &5429 & 2708 & 16 & 4.41 \\
     \hline
    Citeseer &4715 &3312  & 16 & 1.48 \\
     \hline
    PubMed &44338 &19717  & 16 & 3.64  \\
     \hline
    MSAcademic &81894 &18333  & 16 & 1.14 \\
    \bottomrule
	\end{tabular}} 
\end{table*} 

For the small datasets, training spans days 0--7, validation uses day 8, and testing covers days 9--13. For the medium datasets, training spans days 0--13, validation uses day 14, and testing covers days 15--26. For the large datasets, training spans days 0--94, validation uses day 95, and testing covers days 96--162.

\paragraph{Hyperparameter setup.}
Unless otherwise specified, hyperparameters are kept consistent across methods. The local seed batch size is set to 1k for HI datasets, 2k for LI datasets, and 64 for citation networks (except FedPUB). Hop-1 and hop-2 neighbor sampling sizes are both 100 for AML tasks and 10 for citation networks. The communication interval is fixed at $K=32$, and AML tasks are trained for 30k total iterations, while citation networks are trained for 2k iterations, unless otherwise specified. $\gamma$ is tuned in $[0.1, 0.9]$ in steps of $0.2$. $\beta$ is fixed to be $0.9$. Each algorithm is repeated three times. Performance is evaluated using the average macro F1 across all clients.

\section{Additional Experiments}
\label{sec:app:additional}

\paragraph{Attribute inference attacks.}
We study attribute inference attacks (AIA) on HI-Small to evaluate empirical privacy beyond the formal metric-DP guarantee. AIA represents a realistic threat model in federated GNNs, where adversaries may observe partial embeddings and attempt to infer sensitive node attributes. We assume the adversary knows a target node's membership in a client and a subset of nodes and connections from other clients. Using this information together with shared embeddings, the adversary attempts to reconstruct the unknown node features by minimizing the discrepancy between the embedding of a reconstructed feature vector $x'$ and the true embedding:
\begin{equation}
\arg\min_{x'} \|h^{(L)}(x'; B\cup\{x\}) - h^{(L)}(x; B\cup\{x\})\|^2.
\end{equation}
Reconstruction quality is measured by mean squared error (MSE) between true and reconstructed features.

\begin{figure}[h]
    \centering
    \includegraphics[width=0.3\linewidth]{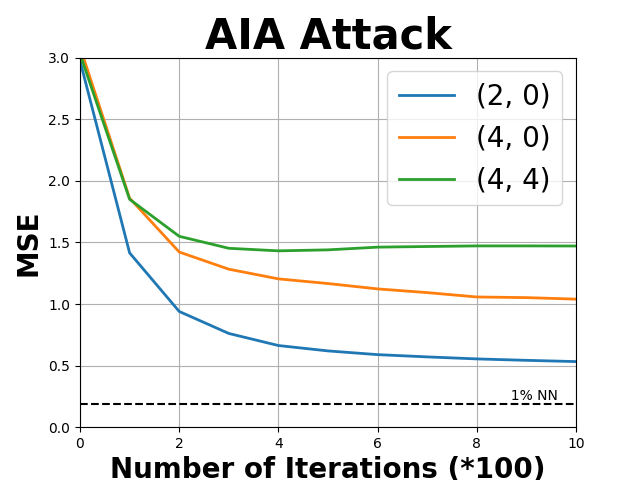}
    \caption{AIA reconstruction error under varying hop-1 and hop-2 neighborhood sizes. Legend entries are in the form of (hop-1, hop-2).}
    \label{fig:aia}
\end{figure}

Figure~\ref{fig:aia} reports MSE under varying hop-1 and hop-2 neighborhood sizes. Embeddings aggregated over larger neighborhoods are more difficult to reconstruct, leading to higher reconstruction error. We also report the average distance to the top $1\%$ nearest neighbors as a baseline; reconstruction error exceeds this threshold across all configurations, suggesting that sharing aggregated embeddings provides increased resistance to attribute inference compared to sharing raw or lightly processed features.

\begin{figure}[h]
    \centering
    \includegraphics[width=0.3\linewidth]{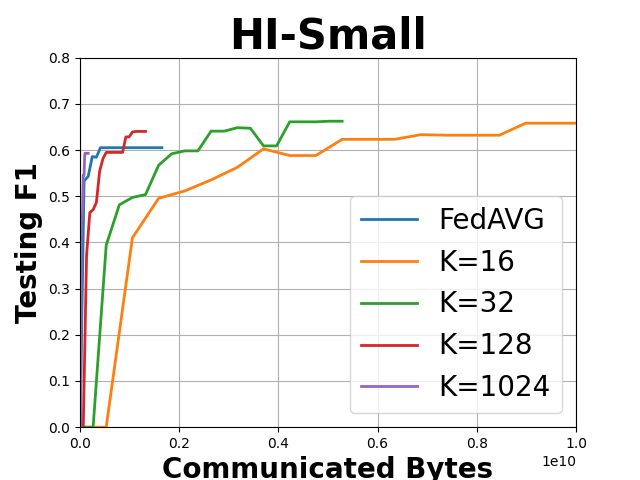}
    \includegraphics[width=0.3\linewidth]{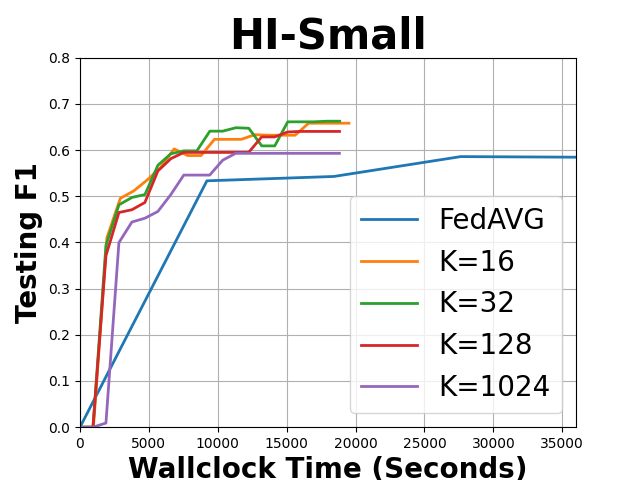}
    \caption{Performance versus communicated bytes/wall-clock time }
    \label{fig:comm_bytes}
\end{figure}

\paragraph{Communication Efficiency.}
Figure \ref{fig:comm_bytes} reports performance as a function of communicated bytes and wall-clock time for our methods compared to the FedAvg baseline. Our methods consistently outperform FedAvg under the same communication budget. Notably, although our approach transmits more bytes per round, it achieves lower wall-clock time at a fixed communication budget, since FedAvg requires substantially more rounds to reach a comparable budget.

Table~\ref{tab:runningtime} reports average runtime per training round with $K=32$ local steps. Each algorithm was run on a cluster connected by InfiniBand, with each machine using a single NVIDIA A100 GPU. The per-round runtime of CE-FedGNN is comparable to all baselines: with the large communication interval substantially reducing communication overhead, local graph computation becomes the dominant cost across all methods. 

\begin{table*}[tbph] 
	\caption{Average runtime per training round (in seconds) with $K=32$ local steps.}\label{tab:runningtime} 
	\centering 
\scalebox{0.8}{\begin{tabular}{l|c|c|c } 
			\toprule 
 & HI-Small & HI-Medium & HI-Large \\
		\hline  
FedAvg-GIN  & 18.79s & 21.12s &  86.68s    \\ 
Swift-GIN  & 23.51s & 34.92s & 93.44s  \\
FedGCN-GIN & 21.06s & 31.15s &  88.53s\\
FedAvg-PNA  & 28.78s & 40.60s & 151.28s   \\ 
Swift-PNA & 30.53s & 44.58s & 167.09s\\
FedGCN-PNA & 29.01s & 41.56s & 158.72s\\
\hline 
CE-FedGNN-GIN  & 22.68s & 32.34s & 91.36s \\
CE-FedGNN-PNA & 29.45s  & 42.77s  &  164.73s \\
\bottomrule 
	\end{tabular} }
\end{table*}

\paragraph{Ablation Study.}
Table~\ref{tab:ablation} presents an ablation of the moving-average components in CE-FedGNN. Removing global embedding sharing causes a substantial drop in performance across all datasets, with the largest degradation on HI-Medium and LI-Medium. Replacing the moving-average global embeddings with stale (non-averaged) versions recovers part of this loss but remains below the full method, particularly on LI-Medium, where representation drift across rounds is more pronounced. Removing the gradient moving-average estimator underperforms the full method on most datasets, except on LI-Medium. 

\begin{table*}[h]
\caption{Ablation study on AML datasets. Best results in bold.}
\label{tab:ablation}
\centering
\scalebox{0.8}{\begin{tabular}{l|c|c|c|c}
\toprule
& HI-Small & HI-Medium & LI-Small & LI-Medium \\
\hline
w/o global embedding & $0.5427\pm0.0288$ & $0.5037\pm0.0341$ & $0.1556\pm0.0131$ & $0.2614\pm0.0225$ \\
Stale global embedding & $0.5436\pm0.0311$ & $0.5382\pm0.0276$ & $0.1614\pm0.0142$ & $0.1129\pm0.0173$ \\
w/o gradient moving average & $0.5666\pm0.0252$ & $0.6457\pm0.0330$ & $0.1883\pm0.0209$ & $\mathbf{0.3535\pm0.0236}$ \\
CE-FedGNN (full) & $\mathbf{0.6623\pm0.0273}$ & $\mathbf{0.6517\pm0.0322}$ & $\mathbf{0.2655\pm0.0199}$ & $0.2918\pm0.0106$ \\
\bottomrule
\end{tabular}}
\end{table*}

\textbf{Metric-DP guarantees.}
We instantiate the metric-DP analysis of Theorem~\ref{thm:mdp_composition} on HI-Small embeddings. Following Section~\ref{sec:privacy}, we $L_2$-normalize released embeddings and report the empirical $k$-th nearest neighbor distance distribution $\{d_k(v) : v \in V\}$ at $k=50$. Table~\ref{tab:metric_dp_hs} and ~\ref{tab:metric_dp_hm} report $\rho_q$ at percentile levels $q \in \{50, 90, 95, 99, 100\}$ together with the corresponding $(\eps, \delta)$-metric-DP guarantee for different $\sigma_0$, $\delta = 10^{-4}$, $T = 30\text{k}$ iterations.

The $90$th-percentile guarantee provides a $k$-anonymity-style claim that holds for at least $90\%$ of accounts in the dataset. The full distribution allows readers to characterize the privacy guarantees for typical accounts and outliers separately. 

\begin{table}[htbp]
\centering
\caption{Empirical $\rho_p$ and corresponding $(\eps,\delta)$-metric-DP guarantees on HI-Small at $k=50$, $\delta=10^{-4}$, $T=30\text{k}$. Headline guarantee is reported at the 90th percentile.}
\begin{small}
\begin{tabular}{l|c|c|c|c|c|c|c}
\toprule
Percentile $q$ & $\sigma_0=0.3$ & $\sigma_0=0.5$ &$\sigma_0=0.7$ &$\sigma_0=1$ &$\sigma_0=2$ &$\sigma_0=3$ &$\sigma_0=5$ \\
\midrule
50th ($\rho_q=0.0533$) & 12.881 & 6.815 & 4.556 & 3.005 & 1.367 & 0.869 & 0.492 \\
90th ($\rho_q=0.1466$) & 51.794 & 25.017 & 15.941 & 10.097 & 4.356 & 2.719 & 1.522 \\
95th ($\rho_q=0.1845$) & 73.245 & 34.476 & 21.641 & 13.524 & 5.724 & 3.547 & 1.973 \\
99th ($\rho_q=0.2793$) & 141.039 & 63.316 & 38.576 & 23.426 & 9.501 & 5.787 & 3.173 \\ 
100th ($\rho_q=0.8913$) & 1059.705 & 424.668 & 237.899 & 131.634 & 45.275 & 25.478 & 12.936 \\
\bottomrule
\end{tabular}
\end{small}
\label{tab:metric_dp_hs}
\end{table}

\begin{table}[htbp]
\centering
\caption{Empirical $\rho_p$ and corresponding $(\eps,\delta)$-metric-DP guarantees on HI-Medium at $k=50$, $\delta=10^{-4}$, $T=30\text{k}$. Headline guarantee is reported at the 90th percentile.}
\begin{small}
\begin{tabular}{l|c|c|c|c|c|c|c}
\toprule
Percentile $q$ & $\sigma_0=0.3$ & $\sigma_0=0.5$ &$\sigma_0=0.7$ &$\sigma_0=1$ &$\sigma_0=2$ &$\sigma_0=3$ &$\sigma_0=5$ \\
\midrule
50th ($\rho_q=0.0339$) & 4.823 & 2.660 & 1.813 & 1.214 & 0.561 & 0.358 & 0.203 \\
90th ($\rho_q=0.1767$) & 41.001 & 20.150 & 12.957 & 8.276 & 3.613 & 2.265 & 1.273 \\
95th ($\rho_q=0.2143$) & 54.408 & 26.194 & 16.652 & 10.523 & 4.530 & 2.825 & 1.580 \\
99th ($\rho_q=0.2988$) & 90.373 & 41.868 & 26.045 & 16.128 & 6.741 & 4.156 & 2.303 \\
100th ($\rho_q=0.7724$) & 441.083 & 183.477 & 106.161 & 61.256 & 22.713 & 13.291 & 7.020 \\
\bottomrule
\end{tabular}
\end{small}
\label{tab:metric_dp_hm}
\end{table}

~
\newpage
~
\newpage

\end{document}